%% file: revised_ccs.tex
\documentclass[sigconf]{acmart}

\AtBeginDocument{%
  \providecommand\BibTeX{{%
    \normalfont B\kern-0.5em{\scshape i\kern-0.25em b}\kern-0.8em\TeX}}}

\copyrightyear{2021}
\acmYear{2021}
\setcopyright{acmcopyright}\acmConference[CCS '21]{Proceedings of the 2021 ACM SIGSAC Conference on Computer and Communications Security}{November 15--19, 2021}{Virtual Event, Republic of Korea}
\acmBooktitle{Proceedings of the 2021 ACM SIGSAC Conference on Computer and Communications Security (CCS '21), November 15--19, 2021, Virtual Event, Republic of Korea}
\acmPrice{15.00}
\acmDOI{10.1145/3460120.3484579}
\acmISBN{978-1-4503-8454-4/21/11}

\usepackage{cite}
\usepackage{soul}
\usepackage{diagbox}
\usepackage{pifont}
\usepackage{graphicx}
\usepackage{textcomp}
\usepackage{xcolor}
\usepackage[utf8]{inputenc} 
\usepackage[T1]{fontenc}    
\usepackage{hyperref}       
\usepackage[nameinlink]{cleveref}
\usepackage{url}            
\usepackage{booktabs}       
\usepackage{nicefrac}       
\usepackage{microtype}      
\usepackage{xspace}
\usepackage{multirow}
\usepackage{multicol}
\usepackage{enumitem}
\usepackage{wrapfig}
\usepackage{color}
\usepackage[normalem]{ulem}
\input{macros}

\DeclareMathOperator\erf{erf}

\newcommand{\cmark}{\ding{51}}%
\newcommand{\xmark}{\ding{55}}%

\begin{document}

\fancyhead{}

\title{DataLens: Scalable Privacy Preserving Training via Gradient Compression and Aggregation}

\author{Boxin Wang}
\authornote{Authors contributed equally to this research.}
\email{boxinw2@illinois.edu}
\affiliation{%
  \institution{University of Illinois at Urbana-Champaign}
    \city{Illinois}
  \country{USA}
}

\author{Fan Wu}
\authornotemark[1]
\email{fanw6@illinois.edu}
\affiliation{%
  \institution{University of Illinois at Urbana-Champaign}
    \city{Illinois}
  \country{USA}
}

\author{Yunhui Long}
\authornotemark[1]
\email{ylong4@illinois.edu}
\affiliation{%
  \institution{University of Illinois at Urbana-Champaign}
    \city{Illinois}
  \country{USA}
}

\author{Luka Rimanic}
\email{luka.rimanic@inf.ethz.ch}
\affiliation{%
  \institution{ETH Zürich}
  \city{Zürich}
  \country{Switzerland}
}

\author{Ce Zhang}
\email{ce.zhang@inf.ethz.ch}
\affiliation{%
  \institution{ETH Zürich}
  \city{Zürich}
  \country{Switzerland}
}

\author{Bo Li}
\email{lbo@illinois.edu}
\affiliation{%
  \institution{University of Illinois at Urbana-Champaign}
  \city{Illinois}
  \country{USA}
}

\begin{abstract}

\input{abstract}
\end{abstract}



\maketitle

\section{Introduction}\label{sec:intro}
\input{introduction}

\section{Preliminaries}
Here we will first provide some background knowledge on differential privacy and data generative models. 
We then draw connections between the definitions we introduced here and our analysis on \sysname later.

\input{preliminaries}

\section{Threat Model $\&$ Method Overview}
In this section, we will first introduce the threat model that we consider in this paper, then provide an overview of the proposed \sysname framework as a differentially private data generative model.
We also provide an overview of the proposed noisy gradient compression and aggregation method \top, which serves as one of the key building blocks in \sysname.

\subsection{Threat Model and Goal}
In practice, the machine learning models are usually trained by data containing a large amount of privacy sensitive information. 
Thus, given a trained model, an attacker is able to train some shadow models with partial data or leverage other strategies to infer the ``membership” of a training instance~\citep{membershipAttack}, which leads to the leakage of sensitive information.
For instance, if a person is known to have participated in a heart disease test, her privacy of having heart disease would be revealed. An attacker is also able to recover the training information via data recovery attacks~\citep{carlini2019secret,carlini2020attack}.

Differential privacy (DP) can protect against \textit{membership inference attacks} and \textit{training-data memorization}~\citep{yeom2018privacy, carlini2019secret}. Intuitively, differential privacy guarantees that when the input dataset differs by one record, the output distribution of a differentially private algorithm does not change by much. This definition reduces the risk of membership inference attacks and data recovery attacks 
given that it prevents the algorithm from memorizing individual record in the input training dataset.

In this paper, our \textbf{goal} is to ensure the differential privacy guarantees for training machine learning models, and therefore protect the privacy of training data. 
There has been a line of research focusing on providing differential privacy guarantees for the trained machine learning models by adding DP noise during training~\citep{Abadi2016DeepLW}.
Here we mainly consider a more flexible case, where we will design a differentially private data generative model, which ensures that the \textit{generated data} instead of the model's parameters are differentially private as proved in~\Cref{theorem:rdp_guarantee}. Thus, as long as the data are generated, they can be used for training arbitrary down-stream learning tasks with differential privacy guarantees.

Note that besides privacy-preserving, it is also critical to make sure that the generated data is of high \textbf{utility}, and therefore we evaluate the \textit{prediction accuracy} of models trained on the DP generated data and test their accuracy on real testset. Different with existing data generative models,  ``visual" quality of the generated DP data is not the main goal of this paper, and we will provide evaluation on the visual quality of the generated data for understanding purpose in~\Cref{sec:main-results} and~\Cref{tab:visual}. We believe it is interesting future research to integrate other losses to further improve the visual quality of the generated data if it is part of the goal.

\subsection{Method Overview}\label{sec:overview}

\begin{figure*}
\centering
\includegraphics[width=0.7\linewidth]{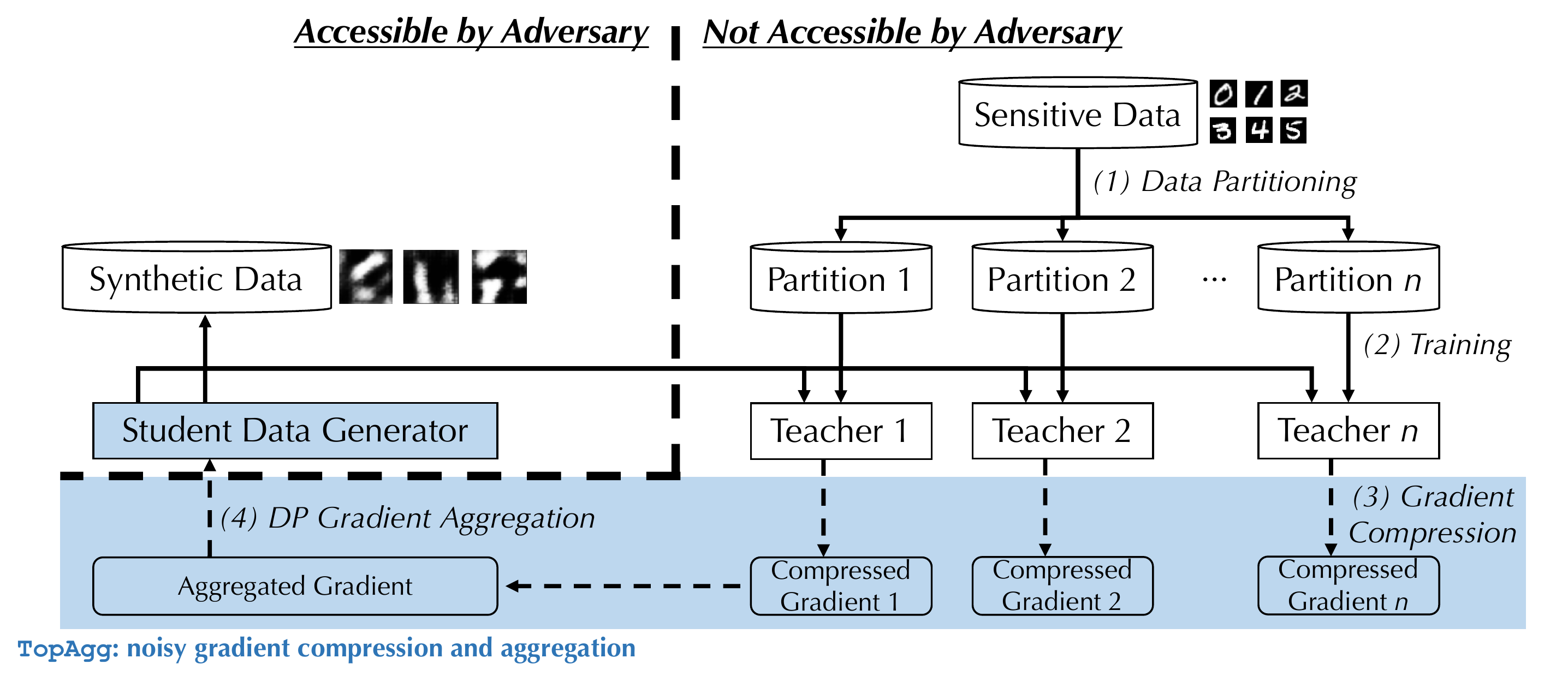}
\vspace{-3mm}
\caption{\small \m{\textbf{Overview of \sysname.} \sysname consists of an ensemble of teacher discriminators and a student generator. \sysname provides a novel algorithm \top for \emph{high dimensional DP gradient compression and aggregation}. \top consists of two parts: (1) top-$k$ and
sign gradient compression that selects the top $k$ gradient dimensions, and (2) DP gradient aggregation for high-dimensional sparse gradients. The solid arrows denote the data flow, while the dash arrows denote the gradient flow.}}
\label{fig:overview}
\end{figure*}

Here we briefly illustrate the proposed \sysname framework, as well as the novel noisy gradient compression and aggregation approach \top which serves as a key building block in \sysname.
The \textbf{goal} of \sysname is to generate high-dimensional data which will not leak private information in the training data. \m{In terms of privacy preserving ML training, PATE~\citep{papernot2018scalable} so far has achieved the state of the art performance, which motivates our privacy analysis.}
\Cref{fig:overview} presents an overview for the structure of \sysname. This framework combines the algorithm \top for high dimensional differentially private (DP) gradient compression and aggregation with GAN and the PATE framework. 
\sysname consists of an ensemble of teacher discriminators and a student generator. 
The teacher discriminators have access to randomly partitioned  non-overlapping sensitive training data. \m{In each \textit{training iteration}, each teacher model produces a gradient vector to guide the student generator in updating its synthetic records. These gradient vectors from different teachers are compressed and aggregated using the proposed DP gradient aggregation algorithm \top before they are sent to the student generator.}

The DP gradient compression and aggregation step is crucial for the privacy protection and utility of the generator. 
Yet, it is challenging for the algorithm to both preserve high data utility and achieve a strong privacy guarantee. 
To achieve high data utility, the algorithm needs to preserve the correct gradient directions of the teacher models. 
As for the privacy guarantee, privacy composition over a high dimensional gradient vector often consumes high privacy budget, resulting in a weaker privacy guarantee. 

To address this problem, prior work uses random projection to project the gradient vector onto lower dimensions~\citep{long2019scalable}. However, this approach introduces excessive noise to the gradient directions and greatly undermines the utility of the model, making it hard to analyze the convergence.

In \sysname, we propose a novel algorithm \top for high dimensional DP gradient compression and aggregation. \m{Our main insight hinges on gradient sparsification
as indicated in recent 
work on communication-efficient distributed learning~\citep{NIPS2017_6768, NIPS2018_7837}: 
we can apply aggressive lossy
compression on the 
gradient vectors without
slowing down SGD convergence.
In this paper, we identify
a specific lossy compression scheme under which we can 
leverage more efficient DP mechanism, thus
increasing the utility significantly.}

In particular, the proposed gradient compression and aggregation algorithm \top takes the top-$k$ entries in a gradient vector and compresses them via stochastic sign gradient quantization~\citep{jin2020stochastic}. This step significantly reduces the dimensionality of a gradient vector while preserving the most valuable gradient direction information. After the compression, we perform DP gradient aggregation over the sign gradient vectors with a corresponding noise injection mechanism. Since the gradient vectors have been compressed, the aggregation algorithm has a much \textit{lower sensitivity}, which leads to a tighter privacy bound. We have also provide a theoretical analysis for the convergence of \top in~\Cref{sec:convergence}, which to our best knowledge is the first convergence analysis considering the \textit{coordinate-wise} gradient clipping together with gradient compression and DP noise mechanism.

\section{\sysname: Scalable Privacy Preserving Generative Model}

We first present our privacy preserving data generative model \sysname, then perform a rigorous analysis on its privacy guarantee and convergence, and demonstrate the privacy-utility trade-off controlled by the proposed gradient compression method. We also briefly discussion how to adapt the proposed noisy gradient compression and aggregation algorithm \top from \sysname to standard SGD training.

\vspace{-0.5em}
\subsection{\sysname Training}
\label{sec:training}

\input{algorithm}

\vspace{-0.8em}
\subsection{Differential Privacy Analysis for \sysname}
\label{sec:dp-analysis}

In this section, we analyze the differential privacy bound for the proposed \sysname framework, and we leverage the R\'enyi differential privacy in our analysis. \m{We also compare the data-dependent privacy bound and the data-independent privacy bound, and  we show the data-independent one  is more suitable for analyzing \sysname. }
\vspace{-0.5em}
\paragraph{R\'enyi Differential Privacy} We utilize R\'enyi Differential Privacy (RDP) to perform the privacy analysis since it supports a tighter composition of privacy budget and can be applied to both data-independent and data-dependent settings. First, we review the definition of RDP and its connection to DP. 

\begin{definition}[$(\lambda, \alpha)$-RDP~\citep{mironov2017renyi}]
\label{def:rdp}
A randomized mechanism $\mathcal{M}$ is said to guarantee $(\lambda, \alpha)$-RDP with $\lambda > 1$ if for any neighboring datasets $D$ and $D'$,
\begin{small}
\begin{equation*}
    D_{\lambda}\left(\mathcal{M}(D) \| \mathcal{M}\left(D^{\prime}\right)\right) = 
    \frac{1}{\lambda-1} \log \mathbb{E}_{x \sim \mathcal{M}(D)}\left[\left(\frac{\mathbf{P} \mathbf{r}[\mathcal{M}(D)=x]}{\mathbf{P r}\left[\mathcal{M}\left(D^{\prime}\right)=x\right]}\right)^{\lambda-1}\right] \leq \alpha.
\end{equation*}
\end{small}
\end{definition}

For any given probability $\delta>0$, $(\lambda, \alpha)$-RDP implies $(\varepsilon_\delta, \delta)$-differential privacy with $\varepsilon_\delta$ bounded by the following theorem. 
\m{The definition of \textit{neighboring dataset} in this work follows the standard definition used in PATE framework~\citep{papernot2016semi} and DP-SGD framework~\citep{Abadi2016DeepLW}. As noted in \citet{Abadi2016DeepLW}, the neighboring datasets would differ in a single entry, that is, one image instance is \textit{present} or \textit{absent} in one dataset compared with the other taking image  as an example.
}

\begin{theorem}[From RDP to DP~\citep{mironov2017renyi}]
\label{theorem:rdp-dp}
If a mechanism $\mathcal{M}$ guarantees $(\lambda, \alpha)$-RDP, then $\mathcal{M}$ guarantees $(\alpha + \frac{\log 1/\delta}{\lambda-1}, \delta)$-differential privacy for any $\delta \in (0,1)$.
\end{theorem}

In the remaining of this section, we first use RDP to analyze the privacy bound of $\sysname$, and then derive the final DP bound in \Cref{theorem:rdp_guarantee}. We will first analyze the data-independent and data-dependent privacy bounds.

\paragraph{Data-Independent Privacy Bound}

\input{data_independent_privacy_bound}

\paragraph{Data-Dependent Privacy Bound}

\input{data_dependent_privacy_bound}
\subsection{Convergence Analysis of \top } 
\label{sec:convergence}
\input{convergence_analysis}

\section{Experimental Evaluation}
In this section, we present the experimental evaluation of \sysname for generating differentially private data with high utility. We compare \sysname with state-of-the-art differentially private generative models and evaluate the data utility and visual quality on high-dimensional image data such as CelebA face and Places365 to demonstrate the effectiveness and scalability of  \sysname.

\vspace{-0.5em}
\input{tables/tab_main}
\subsection{Experimental Setup} 
\label{sec:exp-gen-setup}

We compare the generated data utility of \sysname with three state-of-the-art baselines: DP-GAN~\citep{xie2018differentially}, PATE-GAN~\citep{yoon2018pategan}, GS-WGAN~\citep{chen2020gs}, and G-PATE~\citep{long2019scalable} on four image datasets. 

\vspace{1mm}
\textbf{Datasets.}\quad
To demonstrate the advantage of \sysname as being able to generate high dimensional differentially private data, we focuse on high dimensional image datasets, including MNIST~\citep{lecun1998mnist}, Fashion-MNIST~\citep{xiao2017fashion}, CelebA datasets~\citep{liu2015faceattributes}, and \m{Places365 dataset~\citep{zhou2017places}}. MNIST and Fashion-MNIST dataset contain grayscale images of $28 \times 28$ dimensions. Both datsets have 60,000 training examples and 10,000 testing examples. The CelebA dataset contains 202,599 color images of celebrity faces. We use the official preprocessed version with face alignment and resize the images to $64 \times 64 \times 3$. 
\m{Places365 dataset is consisted of 1.8M high resolution color images of diverse scene categories. We select three level-2 classes to compose a dataset of size 120,000 and resize the images to $64 \times 64 \times 3$.}

We create two CelebA datasets based on different attributes: CelebA-Gender is a binary classification dataset with gender as the label, while CelebA-Hair uses three hair color attributes (black/ blonde/ brown) as classification labels. The training and testing set is split following the official partition as~\citep{liu2015faceattributes}. 
Since DP-GAN and PATE-GAN did not evaluate their framework on high dimensional image datasets, we run their open-source code and compare with the proposed \sysname framework.

\vspace{1mm}
\textbf{Models.}\quad
Both the teacher discriminator and the student generator of \sysname uses the same architecture as DC-GAN~\citep{radford2015unsupervised}. The latent variables sampled from Gaussian distribution are 50-dimensional for MNIST, \m{50-dimensional ($\eps=1$) and} 64-dimensional \m{($\eps=10$)} for Fashion-MNIST, 100-dimensional for CelebA datasets, \m{and 100-dimensional for Places365}. 
For $\eps=1$, 
we set top-$k$=200 for MNIST and Fashion-MNIST, top-$k$=700 for CelebA and Places365.
\m{For $\eps=10$,
we set top-$k$=350 for MNIST and Fashion-MNIST, top-$k$=500 for CelebA, and top-$k$=700 for Places365.}
\m{Ablation studies and discussions on comprehensive hyper-parameter analysis can be found in~\Cref{sec:exp-datalens-result}}.

\textbf{Baselines.}\quad For baseline models, DP-GAN uses standard WGAN and adds Gaussian noise on the gradients during training to achieve differential privacy. Both PATE-GAN and G-Pate leverage PATE framework to generate differentially private images based on different teacher aggregation strategies. Since DP-GAN and PATE-GAN did not evaluate or report their frameworks on (high-dimensional) image datasets, we run their open-source code of DP-GAN\footnote{Code at \url{https://github.com/illidanlab/dpgan}} and PATE-GAN\footnote{Code at \url{https://bitbucket.org/mvdschaar/mlforhealthlabpub/src/master/alg/pategan/}} and compare with our \sysname framework.   \m{For GS-WGAN, we use its open-source implementation\footnote{Code at \url{https://github.com/DingfanChen/GS-WGAN}} to train DP generative models.  
For large $\varepsilon=10$, we can reproduce the performance on MNIST and Fashion-MNIST.
Under small $\varepsilon=1$ setting, we tried our best to tune the hyper-parameters of GS-WGAN; however, we observe GS-WGAN is unable to converge given the limited privacy constraints, especially when presented with higher-dimensional data (CelebA, Places365) which is confirmed with the authors. }

\textbf{Evaluation Metrics.}\quad 
We follow standard evaluation pipelines \citep{long2019scalable, yoon2018pategan, chen2020gs} and evaluate \sysname as well as baselines \m{in terms of \textit{data utility} and \textit{visual quality} under different privacy constraints. Specifically, \textit{data utility} is evaluated by training a classifier with the generated data and testing the classifier on real test dataset. We consider the \textit{testing accuracy} on the test set as the indicator for the utility of the synthetic data for downstream tasks. } 
\m{To evaluate the visual quality of generated data for understanding purpose, we consider \textit{Inception Score (IS)} \citep{li2017alice} and \textit{Frechet Inception Distance (FID)} \citep{salimans2016improved}, which are standard metrics of visual quality in GAN literature.} 
We also provide the images generated by \sysname in Appendix~\Cref{fig:vis} for visualization.

\vspace{-0.5em}
\subsection{Experimental Results}
\label{sec:main-results}
\label{sec:exp-gen-results}
In this section, we evaluate \sysname on different datasets. We first compare the generated data utility  for \sysname and four other state of the art DP generative model baselines.
We then explore  the performance of \sysname under limited privacy budgets (\ie, $\epsilon <1$), which is a  challenging while important scenario. 
\m{We then evaluate the visual quality of the generated data, followed by a range of ablation studies on the data-dependent and data-independent privacy analysis, 
impacts of different hyper-parameters and components in \sysname, as well as different compression methods. 
We show that the proposed \sysname not only outperforms all baselines, but also demonstrates additional advantages especially when the privacy budget is small.}

\input{tables/tab_main_small}

\indent {\textbf{Data Utility Evaluation.}}\quad
We first compare \sysname with \m{four} baselines
under two privacy budget settings $\varepsilon=1, \delta=10^{-5}$ and  $\varepsilon=10, \delta=10^{-5}$ on \m{five} high dimensional image datasets,
following the standard evaluation pipeline. 

From~\Cref{tab:image}, we can see that \sysname shows substantially higher performance than all baseline methods especially when $\varepsilon=1$. In particular, the performance improvement on MNIST under $\varepsilon=1$ is more than $13\%$. 
Even for high dimensional datasets like CelebA-Hair and \m{Places365} whose dimensionality is 16 times larger than MNIST, \sysname achieves $10\%$ higher performance improvement than the state of the art, which demonstrates its advantages on high dimensional data than other baseline DP generative models. 
\m{Specifically, we note that GS-WGAN can only converge under large privacy budget ($\varepsilon=10$) for gray-scale datasets (MNIST and FashionMNIST), as GS-WGAN needs 20k epochs and small noise to converge. 
In comparison, \sysname can converge within 100 epochs due to the fast convergence rate brought by top-$k$ operation for high-dimensional datasets.
As a result, under the limited privacy budget ($\varepsilon=1$) or given high dimensional facial datasets (\eg, CelebA), GS-WGAN is unable to converge and therefore generate low-utility data, making the classifier accuracy close to random guessing; while \sysname can generate high-utility data even with limited privacy budget.
}

\textbf{Evaluation under small privacy budget.}\quad
To further demonstrate the advantage of \sysname as being able to generate high-utility images under \textit{small} privacy budgets (\ie, higher privacy protection guarantees), we conduct ablation studies on MNIST and Fashion-MNIST under $\varepsilon \le 1$. The experimental results are shown in~\Cref{tab:lowdp}.

We find that \sysname achieves the best results compared with the baselines given such tight privacy constraints.  With increasing privacy budgets, different DP models gradually converge and the accuracy increases. We note that \sysname converges the fastest and achieves more than $20\%$ accuracy even under smallest privacy budget $\varepsilon=0.2$ for both MNIST and Fashion-MNIST datasets; while the baseline models barely converge and the accuracy is similar to random guess. The observed experimental results also support our theoretical analysis that the proposed \top algorithm can introduce a smaller bias and provide high-utility gradient information for the student generator to converge, demonstrating that our method is particularly effective under limited privacy budgets.

\m{{\textbf{Visual Quality Evaluation.}}\quad} \m{We present the quantitative visual quality evaluation of \sysname and baselines in Table \ref{tab:visual} based on Inception Score (IS) under different privacy constraints:  $\varepsilon=1, \delta=10^{-5}$ and  $\varepsilon=10, \delta=10^{-5}$. Since CelebA-gender and CelebA-face are from the same distribution of face images and have a lot of overlapping, we mainly consider CelebA-Gender  to represent the visual quality for CelebA face dataset.}

\m{We observe that \sysname consistently outperform the baselines in terms of visual quality while ensuring the rigorous privacy protection when $\varepsilon=1$, which suggests that \sysname can converge faster than the state-of-the-art baselines. Specifically, the generated differentially private MNIST images achieve the inception score of 4.37, improving the strongest baseline G-PATE by more than $20\%$. When $\varepsilon=10,$ we find that \sysname can be outperformed by GS-WGAN  on MNSIT and Fashion-MNIST, but still outperforms all baselines on high-dimensional CelebA datasets. We believe the reason is that while top-$k$ operation can help with faster converge the most important information and yield high-utility data, it may lose some detailed and trivial gradient information for image reconstruction. 
We note that visual quality and data utility are two orthogonal metrics, and \sysname  consistently generates data with the highest utility. We provide the evaluation of FID in Appendix \Cref{tab:full_image}, given that FID is evaluated based on models trained with ImageNet which may not be suitable for evaluating datasets such as MNIST. 
In addition, we believe it would be an interesting future direction to add additional loss terms for improving the visual quality of  the generated data with privacy guarantees.
}

\vspace{-0.5em}
\subsection{Ablation Studies} \label{sec:ablation}
\m{In this section, we conduct a series of ablation studies to further understand the improvements of \sysname, including the empirical exploration of the data-dependent and data-independent privacy bounds, the hyper-parameter impacts, the comparison with different gradient compression methods, as well as the impacts of each component in \sysname pipeline.}

\input{tables/tab_visual}

\m{\textbf{Data-Independent Bound v.s. Data-Dependent Bound.}\quad}
\m{We compute the data-independent privacy bound and the data dependent privacy bound to validate the theoretical comparison in \Cref{sec:dp-analysis}. \Cref{fig:dept_bound}~presents the privacy budget consumption over each training epoch computed by the data-independent bound and the data-dependent bound, respectively. We set $\sigma=5000$ for MNIST and Fashion-MNIST, and $\sigma=9000$ for CelebA-Hair and CelebA-Gender. The training is stopped when the privacy budget $\varepsilon$ computed by the data \-independent bound reaches 1. As shown in \Cref{fig:dept_bound}, the data-independent bound is always tighter than the data-dependent one on the high-dimensional datasets. We also notice that models on MNIST and Fashion-MNIST have a similar data-dependent bound, and so are models on CelebA-Hair and CelebA-Gender. These results align with our theoretical analysis in~\Cref{theorem:rdp-dependent} and \Cref{theorem:data_dependent_prob}. Due to the high dimensionality of the gradients and the Gaussian noise, there is unlikely to be a spike in the probability distribution over the likely outcomes of the gradient aggregation step. Consequently, the data-dependent privacy bound is loose and mostly determined by the dimension of the gradients.  }

\input{tables/fig_dp_bound_main}

\input{tables/tab_hyperpara_all}

\vspace{0.5em}
\textbf{Ablation studies on hyper-parameters.}\quad
\label{sec:exp-datalens-result}
As we can see, \sysname contains several hyper-parameters: the number of the teacher models, the top-$k$, the threshold $\beta$, the standard deviation $\sigma$ of injected Gaussian noise, and the gradient clipping constant $c$. We evaluate a set of hyper-parameters as shown in~\Cref{tab:hyperparam}. For other parameters, we use: for MNIST and Fashion-MNIST datasets, we set $\sigma=5000$ when $\varepsilon=1$ and $\sigma=900$ when $\varepsilon=10$; for CelebA datasets of higher dimensionality, we set $\sigma=9000$ when $\varepsilon=1$ and $\sigma=700$ when $\varepsilon=10$. We set the gradient clipping constant $c=10^{-5}$ in all experiments. We also follow the default DC-GAN model configuration\footnote{Details can be found at \url{https://github.com/carpedm20/DCGAN-tensorflow}.} and set the batch size the same as the disjoint data partition size. 

From~\Cref{tab:hyperparam}, we observe that training more teacher discriminators will give us better performance in general, as it can save more privacy budgets. However, with more teacher discriminators, each discriminator will have access to a smaller amount of training data, thus leading to a slightly worse performance. We observe this trade-off on the CelebA-Gender dataset, where the optimal number of teachers is $6000$. Choosing a proper top-$k$ and $\beta$ is bit tricky: as stated in the discussion of~\Cref{sec:discussion}, if top-$k$ is too small, the model converges slower and is likely to converge to a bad solution. On the other hand, if top-$k$ is too large, we will introduce a larger DP noise and the model can soon reach the privacy budget limit given the high sensitivity. In this paper, we search for the best top-$k$ via grid search. Another observation is that we usually need to set a high threshold $\beta$ to smooth out the noisy gradient entries from the DP noise, though if the threshold is too high it is likely to ignore the top-$k$ voted gradients information. This tradeoff leads us to choose a threshold $\beta$ between $\frac{\sigma}{2N}$ and $\frac{\sigma}{N}$. \m{Finally, we also note that the clipping value $c$ has a large impact on the model convergence. We observe given a fixed top-$k$, a reasonably smaller $c$ generally yields better convergence rate and data utility. This is aligned with our theorem, because if $c$ gets smaller, the convergence bias from the first term $c(d-k)M$ will get smaller. }

\m{We note that the peformance improvements from \sysname does not necessarily comes from the fact that we have more hyper-parameters, since compared to other baseline methods using PATE framework such as G-PATE and PATE-GAN, \sysname only introduces one more hyper-parameter top-$k$ for gradient compression. Moreover, as shown in \Cref{tab:hyperparam}, \sysname can outperform all the baselines on CelebA datasets over a wide range of different hyper-parameters in practice. }

\input{tables/tab_compression}

\m{\textbf{Ablation Studies on the Gradient Compression Methods.}\quad}
Here we analyze the impact of our top-$k$ gradient compression method in \top  compared with other compression methods in previous works, \eg, D$^2$P-\textsc{Fed} and FetchSGD. 

\mo{
In particular, for \dtwop, we replace our~\Cref{algo:topk} (\texttt{TopkStoSignGrad}) that uses stochastic sign compression with it, which essentially uses k-level gradient quantization and random rotation for gradient pre-processing. The detailed algorithm is shown in~\Cref{algo:d2p} and~\Cref{algo:klevel} in~\Cref{sec:app-compression}.
For \fetch, we uses the same stochastic sign compression as we leverage sign signal as teacher voting in PATE framework. During aggregation, we use Count Sketch data structure, and use top-$k$ and unsketch operation to retrieve the aggregated gradient. The detailed algorithm is shown in~\Cref{algo:fetch} in~\Cref{sec:app-compression}.}
From~\Cref{tab:compression}, we note that \dtwop and \fetch are outperformed by our \top in terms of data utility, which is mainly due to the increase of the consumption of privacy budget and  the introduction of additional noise during aggregation.
Concretely, D$^2$P-\textsc{Fed} uses $m$-level gradient quantization, which increases the sensitivity of quantized gradients from $2\sqrt{k}$ to $m\sqrt{k}$. Without top-$k$ mechanism, D$^2$P-\textsc{Fed} compression quickly reaches the limit of the privacy budget, and thus the model barely converges. 
Although FetchSGD uses a similar top-$k$ mechanism during compression, the adoption of Count Sketch data structure introduces additional noisy information when approximating the aggregated gradient, and therefore hurts the utility of the generated data.

\mo{Moreover, we record the running time of \sysname and adapted gradient compression methods \dtwop and \fetch on one Tesla T4 GPU under the best parameters of MNIST, Fashion-MNIST, CelebA-Hair, and Celeb-Gender. The average running for each epoch is shown in Table \ref{tab:time}. The time consumption for \dtwop is significantly higher than \top due to the k-level quantization step as well as the rotation step for gradient transformation. The time consumption for \fetch is significantly higher than \top due to the Count Sketch data structure overhead.}

\mo{
\textbf{Runtime Analysis.}\quad
We record the running time of our framework on one RTX-2080 Ti GPU under the best parameter (4000 teacher) of MNIST for $\varepsilon=1$ for three runs. We then only change the number of teacher discrinimators to 2000 and record the running time again.  The average running time for each epoch given different teacher discriminators are 149.92s for 2000 teachers and 322.17s for 4000 teachers, respectively. The student generator converges within 100 epochs, thus the total training time is around $4-8$ hours for MNIST under $\varepsilon=1$.   The runtime scales almost \textbf{linear} to the number of teachers, so adopting a larger number of teachers will not bring much computation overhead. In contrast, the average training time for DP-GAN and G-PATE takes around $26-34$ hours for MNIST under $\varepsilon=1$. Moreover, GS-WGAN requires hundreds of GPU hours to pretrain one thousand non-private GAN as the warm-up steps. 
}

\begin{table}[t]
\centering
\caption{\small \textbf{Running Time Comparison of different gradient compression methods (\top, D$^2$P-\textsc{Fed}, FetchSGD).}  We report the average training time per epoch on different datasets under $\eps=1$ and $\delta=10^{-5}$.}
\label{tab:time}
\vspace{-2mm}
\tabcolsep=0.11cm
\resizebox{0.8\columnwidth}{!}{
\m{
\begin{tabular}{l|ccc}
\toprule
\diagbox[width=10em]{\textbf{Dataset}}{\textbf{Methods}} & {\textbf{\top}}  & \textbf{D$^2$P-\textsc{Fed}}  & {\textbf{FetchSGD}}     \\
\midrule
{\textbf{MNIST}} & {338.34 s} & {492.43s} & 785.34 s   \\ 
\midrule
{\textbf{Fashion-MNIST}} & \textbf{340.84s} & {471.02s} & 775.35s   \\ 
\midrule
{\textbf{CelebA-Gender}} & \textbf{1196.60s}  & 3683.22s &  2622.40s  \\ 
\midrule
{\textbf{CelebA-Hair}} & \textbf{1120.59s} & 8092.50 s &  2620.63s \\ 
\bottomrule
\end{tabular}%
}
}

\end{table}
\input{tables/tab_components}
\m{\textbf{Ablation Studies on the Impact of Different Components in \sysname.}\quad}  \m{To further understand where the improvements of \sysname come from, we investigate how each component in \sysname pipeline contributes to the generated data utility improvement in Table \ref{tab:components} on four high-dimensional image datasets.}

\m{In particular, we consider the following components: (1) top-$k$, (2) stochastic gradient quantization, and (3) gradient thresholding, and evaluate how they impact the data utility by adding or removing each component. We note that the top-$k$ procedure is the most important component based on results in \Cref{tab:components}, since removing this step will largely increase the privacy consumption, leading models fail to converge when given limited privacy budget. Gradient quantization and thresholding are also useful techniques though less critical, contributing to the $3\% - 7\%$ of the utility improvement as shown in \Cref{tab:components}.}

\section{Related Work}
\label{sec:related}
\input{related_work}

\input{conclusions}

\bibliography{refs}
\bibliographystyle{ACM-Reference-Format}
 
\appendix
\section{\top for DP SGD Training}\label{sec:appen-top}
\subsection{DP SGD Training Algorithm with \top}
\label{sec:sgd-method}

We show the details of the DP SGD training using the \top approach in \Cref{algo:dp-sgd}. 
We mainly adopt the DP SGD framework as in~\citet{Abadi2016DeepLW}.
In particular, at each step of the SGD, we compute the gradient for a random subset of examples, and then use the clipping norm to clip each gradient by their $\ell_2$ norm.
After performing a top-$k$ compression of the gradients to select the sub-dimensions, we then take a sum of the compressed gradients, to which we inject the Gaussian noise subsequently.
Finally, we update the model with the compressed DP gradient.
Theoretically, since the $\ell_2$ norm of the gradient vector is reduced after the top-$k$ compression step, the amount of noise required to achieve the same level of DP guarantee becomes smaller, 
which implies a potentially better utility of the training algorithm.

Note that directly applying the \texttt{TopkStoSignGrad} algorithm (\Cref{algo:topk}) to SGD training does not yield a good utility due to information loss during gradient quantization. 
To overcome this problem, we specially adapt \top to select and preserve a subset of the dimensions in the gradients based on the requirement imposed on the $\ell_2$ norm.
This new strategy is described as the function \texttt{NormTopK} in \Cref{algo:dp-sgd}.
Concretely, 
given a gradient vector $\mathbf{g}$, we compute $\Bar{\bf g}$ by selecting several dimensions in $\mathbf{g}$ with the highest absolute values to ensure that their squared sum is close to the target norm $k \norm{\bf g}^2$ (here $0< k<1$) after the top-$k$ compression, and preserving only the values in the selected dimensions. Thus, the compressed gradient $\Bar{\bf g}$ satisfies the condition $\norm{\Bar{\bf g}}^2 \leq k \norm{\bf g}^2$. The remaining dimensions in $\Bar{\bf g}$ are set to 0 since they contain less information.
In this way, we achieve the goal of gradient compression without suffering a significant distortion.
\m{We note that the $\ell_2$ sensitivity of the gradient sum $\sum_i  \mathbf{\hat g}_t({\bf x}_i)$  is $\sqrt{k} C$ when adapting \top for SGD training, since adding or removing one instance $\bf x$ in the training set would lead to the gradient sum differing by $\hat{\bf g}_t(\bf x)$, whose $\ell_2$ norm is bounded by $\sqrt{k} C$ due to the operations in \texttt{NormTopK}. Thus, the variance of the added Gaussian noise is $kC^2\sigma^2\bf I$.}

\begin{algorithm}[t]
\caption{
\textbf{- Differentially Private SGD training via Gradient Compression and Aggregation \top }
} 
\label{algo:dp-sgd}
\begin{algorithmic}[1]
    \State {\bfseries Input:} 
    Examples \{$\mathbf{x}_1,\ldots,\mathbf{x}_n$\}, Top-$k$-Portion parameter $k$, loss function $\mathcal L(\theta)=\frac{1}{n}\sum_i\mathcal L(\theta, \mathbf{x}_i)$. Parameters: batch size $B$, learning rate $\gamma_t$, noise scale $\sigma$, gradient clipping norm $C$, total number of epochs $T$.
    \Function{\texttt{NormTopK}}{$\bf g,k$}
        \State $norm\leftarrow \norm{\bf g}^2$
        \State $target\leftarrow norm\cdot k$   \hfill\Comment{target norm after processing}
        \State $indices\leftarrow$ pick the dimensions in a decreasing order in terms of the squared norm at that dimension, so that the sum of the squared norms in those dimensions add up to right below $target$.
        \State $\mathbf{\tilde g}\leftarrow$ preserve the values of $\mathbf{g}$ at $indices$ and set value at other dimensions as $0$
        \State\Return $\mathbf{\tilde g}$
    \EndFunction   
    \State
    \State Initialize $\theta_0$ randomly
    \For{epoch $t \in [T]$}
        \State Sample a batch of instances $\{\mathbf{x}_{t_i}\}_{i=1}^{B}$ each with sampling probability $B/n$.
        \For{each sample $\mathbf{x}\in\{\mathbf{x}_{t_i}\}_{i=1}^{B}$}
            \State ${\bf g}_t(\mathbf{x})\leftarrow \nabla_{\theta_t} \mathcal L(\theta_t, \mathbf{x})$\hfill\Comment{compute gradient}
            \State ${\mathbf{\tilde g}}_t(\mathbf{x})\leftarrow \bf g_t(x)/\rm{max}\left(1,\frac{\norm{\bf g_t(\mathbf{x})}}{C}\right)$ \hfill\Comment{clip gradient}
            \State $\mathbf{\hat g}_t(\mathbf{x})\leftarrow \mathtt{NormTopK}({\mathbf{\tilde g}}_t(\mathbf{x}), k)$ \hfill\Comment{compress gradient}
        \EndFor
        \State $\mathbf{\bar g}_t\leftarrow \frac{1}{B}\left(\sum_i  \mathbf{\hat g}_t(\mathbf{x}_i) +\mathcal N(0,k\sigma^2C^2\bf I)\right)$ \hfill\Comment{add noise}
        \State $\theta_{t+1}\leftarrow \theta_t-\gamma_t \mathbf{\bar g}_t$ \hfill\Comment{gradient descent}
    \EndFor
    \State {\bfseries Output} $\theta_t$ and compute the overall privacy cost $(\eps, \delta)$ using Moments Accountant
\end{algorithmic}
\end{algorithm}

\subsection{Evaluation of \top for DP SGD}
\label{sec:exp-sgd-setup}

In this section, we demonstrate the universality of the proposed DP gradient compression and aggregation algorithm \top in \sysname, and in particular, the feasibility of applying it to DP SGD training by 
evaluating its performance on two standard image classification tasks for evaluating DP SGD mechanisms.
We first describe the experimental setup.
Then we provide extensive evaluation results on a wide range of privacy budgets. 
Overall, the results show that the \top enabled DP SGD training achieves similar or even better performance on model utility compared with the state-of-the-art Gaussian DP mechanism based on the Moment Account~\citep{mcmahan2019tfprivacy} (we will call it \gmdp in the rest of the paper). We also show that \top would bring additional advantages  under limited privacy budgets, where the utility gap between the DP model and the vanilla model is large for traditional DP approaches.

\subsubsection{Experimental Setup} \hfill \\
We evalute \top for DP SGD training on two datasets, and compare its performance with two baseline frameworks, including the differentially private deep learning (DPDL)~\citep{Abadi2016DeepLW} and \gmdp~\citep{mcmahan2019tfprivacy}.

\textbf{Datasets.}\quad
We experiment with two datasets commonly used in DP SGD research: MNIST~\citep{lecun1998mnist} and CIFAR-10~\citep{cifar10}.
Both datasets are standard image classification datasets. 
The description of MNIST is provided in~\Cref{sec:exp-gen-setup}. Similarly, we use 60,000 instances for training and 10,000 for testing.
CIFAR-10 consists of 60,000 32$\times$32 colored images of 10 classes. Among all, 50,000 instances are used in training and 10,000 are used in testing.

\textbf{Models.}\quad
For MNIST, we adopt a simple convolutional neural network following the default model architecture 
provided in the example in the open source Opacus library\footnote{Code at \url{https://github.com/pytorch/opacus/blob/master/examples/mnist.py}}.
The network is consisted of 
two convolutional layers each followed by a max pooling layer, as well as two fully connected layers on the top.

For CIFAR-10, we follow the setting of DPDL, where we first pretrain the classifiers on public datasets, then freeze the parameters of feature extractor and  finetune on the fully connected layers. In our paper, we use ResNet-18 \citep{he2016deep} as the architecture of the classifier and load the model parameters pretrained on ImageNet\footnote{Publicly available at \url{https://pytorch.org/docs/stable/torchvision/models.html}}. We replace the fully connected layer with a randomly initialized linear head that takes features of $512$-dimension extracted from ResNet feature extractor as input and outputs $10$-dimensional prediction logits. During training, we freeze the parameters of ResNet feature extractor including the parameters of Batch Normalization layers to ensure that the feature extractor will not leak any privacy-sensitive data information.

We compare the performance of \top with two state of the art DP SGD mechanisms: 
\gmdp~\citep{mcmahan2019tfprivacy} and DPDL~\citep{Abadi2016DeepLW}.
In particular, we build upon Opacus~\citep{opacus2020}, a PyTorch implementation of \gmdp that implements the DP SGD training scheme and privacy accountant method in~\citep{mcmahan2019tfprivacy}, which enables  convenient control of randomness in the framework.
Our implementation of \top for DP SGD training is also  built upon the Opacus library.
For DPDL which is the first work that proposed and evaluated the DP SGD training scheme, we directly compare with the  results reported in Section 5.2 and Section 5.3 of the paper for fairness, which present the best model utility performance.

\vspace{1mm}
\textbf{Evaluation Metrics.}\quad
\label{sec:dpsgd-exp}
We adopt \textit{model utility}, which is calculated as the \textit{classification accuracy} of the trained models,  as the evaluation metric for assessing the effectiveness of our algorithm \top. 
For each dataset, each privacy budget $\eps$, and each \texttt{NormTopK} parameter $k$, we perform an extensive grid search for the combination of hyper-parameters (including gradient clipping norm $C$, noise scale $\sigma$, batch size $B$, and learning rate $lr$) for all methods for fair comparison.
We then use the best hyper-parameters to start 10 runs with different random seeds for noise generation and report the averaged results for each method. 
For models trained under baseline frameworks, we follow the same parameter search protocol and parameter grid to obtain 
the reported results. 

\vspace{-0.4em}
\subsubsection{Experimental Results} \hfill \\
\vspace{-0.4em}

\begin{figure*}[h]
    \centering
    \begin{subfigure}[t]{.325\linewidth}
        \centering
        \includegraphics[width=\linewidth]{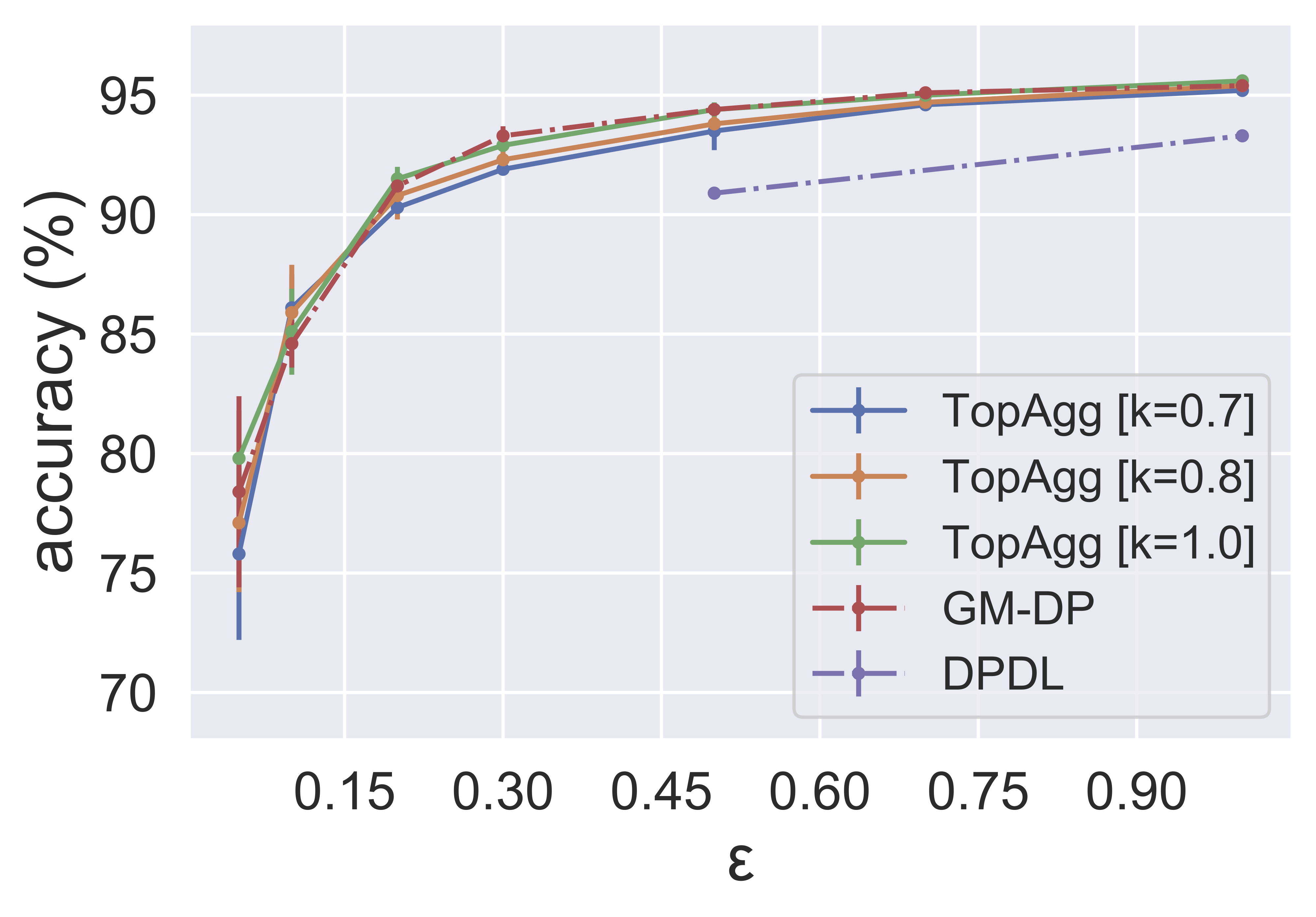}
        \caption{MNIST}
    \end{subfigure}
    \begin{subfigure}[t]{.325\linewidth}
        \centering
        \includegraphics[width=\linewidth]{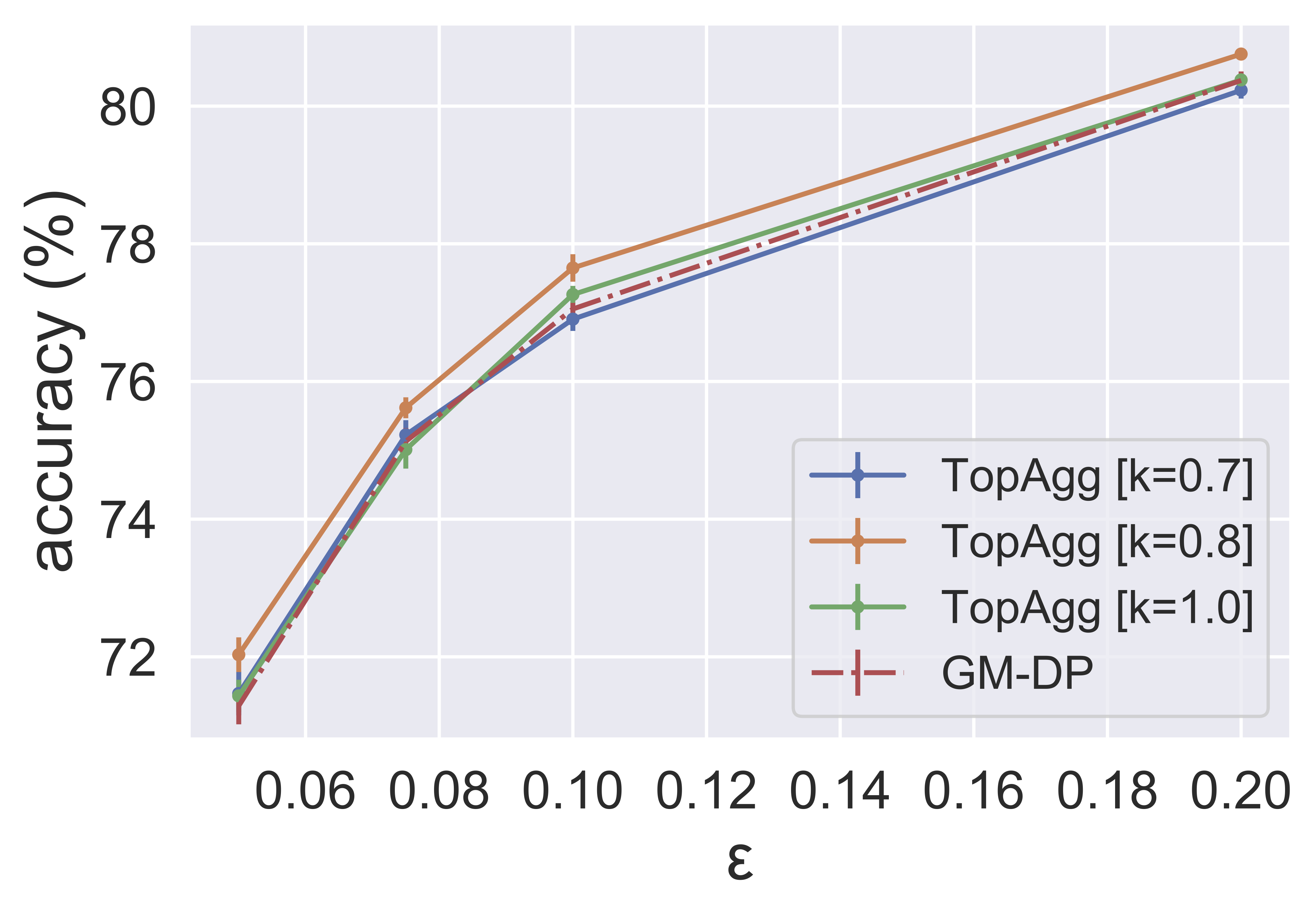}
        \caption{CIFAR-10 (under small $\varepsilon$)}
    \end{subfigure}
        \begin{subfigure}[t]{.325\linewidth}
        \centering
        \includegraphics[width=\linewidth]{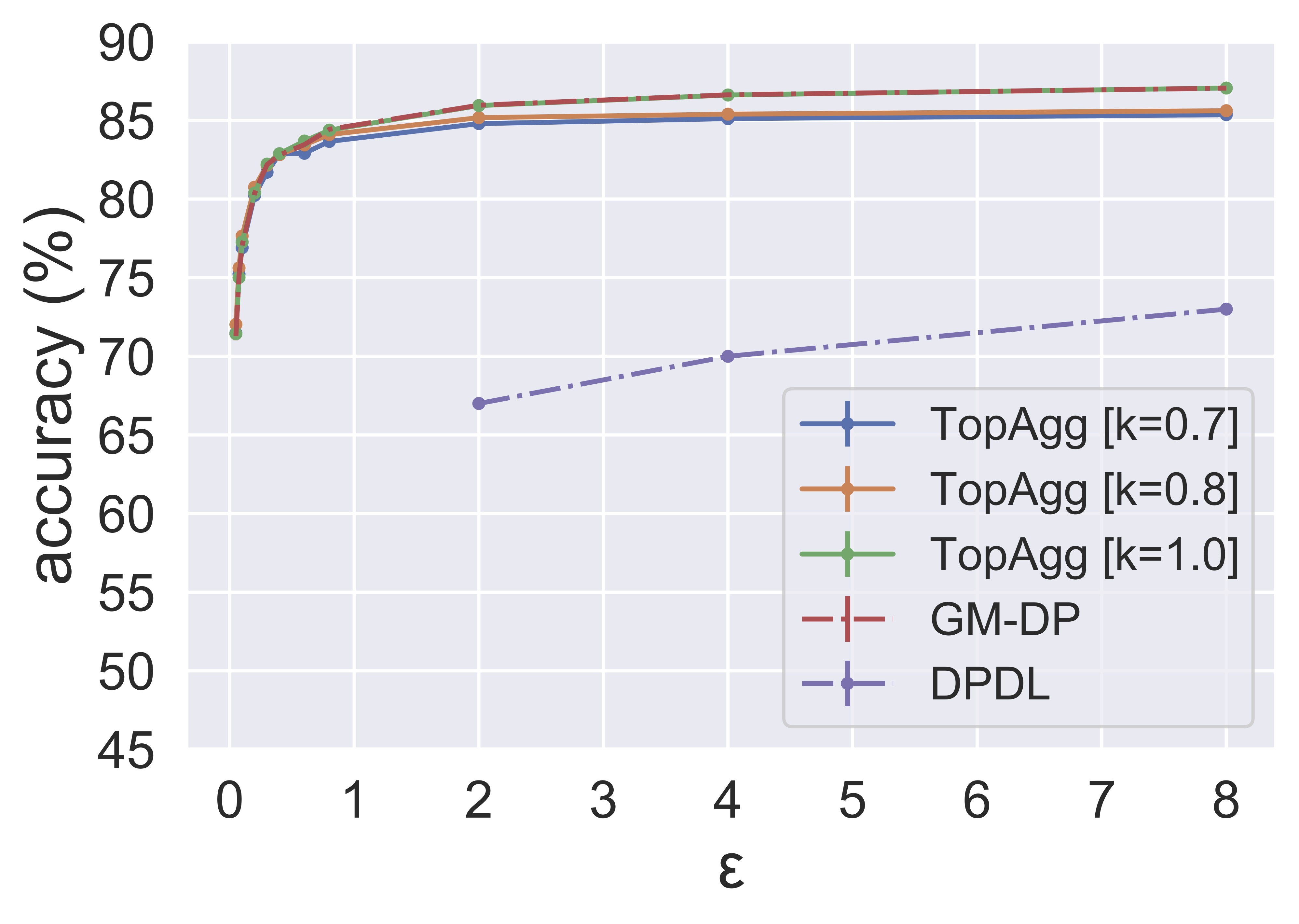}
        \caption{CIFAR-10 (under full range of $\varepsilon$)}
    \end{subfigure}
    \caption{\small The performance of \sysname applied to SGD training on two datasets: (a) MNIST and (b-c) CIFAR-10.
    We run \sysname with a range of $k\in\{0.7,0.8,1.0\}$ and compare its performance with two baselines GM-DP 
    and DPDL on a wide range of privacy budget $\eps$. 
}
    \label{fig:dpsgd}
\end{figure*}

\begin{table}[t]
\centering
\caption{\small Model utility when adapting \top to DP SGD training on (a) MNIST and (b) CIFAR-10 with different privacy parameter $\eps$ and \top parameter $k$. In all cases, $\delta=10^{-5}.$ 
}\label{tab:dpsgd}

\begin{subtable}[]{\columnwidth}
	\centering
	\caption{\small MNIST}\label{tab:dpsgd-mnist}
    \begin{tabular}{l|cccc}
    \toprule
        & \multicolumn{3}{c}{\small \textbf{\top}} & \multirow{2}{*}{\small \textbf{\gmdp}}  \\
    & \multicolumn{1}{c}{\small $k=0.6$}  & \multicolumn{1}{c}{\small $k=0.7$} & \multicolumn{1}{c}{\small $k=0.8$} &   \\\midrule
    $\eps=0.05$      &  73.87 {\tiny  $\pm$  4.77} &  75.81 {\tiny  $\pm$  3.59} &  77.15 {\tiny  $\pm$  2.89}           &  78.40 {\tiny  $\pm$  4.00}\\
    $\eps=0.10$     &  85.67 {\tiny  $\pm$  1.48} &  \textbf{86.12} {\tiny  $\pm$  1.39} &  85.88 {\tiny  $\pm$  1.97}    &  84.55 {\tiny  $\pm$  0.98}\\
    $\eps=0.20$     &  89.84 {\tiny  $\pm$  0.64} &  90.25 {\tiny  $\pm$  0.40} &  90.80 {\tiny  $\pm$  1.04}               &  91.17 {\tiny  $\pm$  0.37}\\
    $\eps=0.30$     &  91.21 {\tiny  $\pm$  0.53} &  91.85 {\tiny  $\pm$  0.21} &  92.35 {\tiny  $\pm$  0.53}              &  \textbf{93.31} {\tiny  $\pm$  0.42}\\
    $\eps=0.50$     &  92.83 {\tiny  $\pm$  0.52} &  93.50 {\tiny  $\pm$  0.75} &  93.82 {\tiny  $\pm$  0.46}              &  \textbf{94.38} {\tiny  $\pm$  0.24}\\
    $\eps=0.70$     &  94.41 {\tiny  $\pm$  0.19} &  94.61 {\tiny  $\pm$  0.24} &  94.65 {\tiny  $\pm$  0.27}              &  \textbf{95.08} {\tiny  $\pm$  0.10}\\
    $\eps=1.00$     &  95.13 {\tiny  $\pm$  0.33} &  95.22 {\tiny  $\pm$  0.20} &  \textbf{95.42} {\tiny  $\pm$  0.20}               &  95.41 {\tiny  $\pm$  0.26}\\
    $\eps=\infty$     &  97.94 {\tiny  $\pm$  0.19} &  98.58 {\tiny  $\pm$  0.11} &  98.79 {\tiny  $\pm$  0.09}          &  \textbf{99.08} {\tiny  $\pm$  0.04}\\
    \bottomrule
    \end{tabular}
\end{subtable}

\begin{subtable}[]{\columnwidth}
	\centering
	\vspace{0.5em}
	\caption{\small CIFAR-10}\label{tab:dpsgd-cifar}
    \begin{tabular}{l|cccc}
    \toprule
    & \multicolumn{3}{c}{\small \textbf{\top}} & \multirow{2}{*}{\small \textbf{\gmdp}}  \\
			& \multicolumn{1}{c}{\small $k=0.6$}  & \multicolumn{1}{c}{\small $k=0.7$} & \multicolumn{1}{c}{\small $k=0.8$}     &   \\\midrule
 $\eps=0.025$      &  42.18 {\tiny  $\pm$  0.89} &  44.32 {\tiny  $\pm$  0.59} &  \textbf{45.87} {\tiny  $\pm$  0.97}   &  41.58 {\tiny  $\pm$  2.01}\\
 $\eps=0.050$      &  65.10 {\tiny  $\pm$  0.48} &  71.47 {\tiny  $\pm$  0.32} &  \textbf{72.03} {\tiny  $\pm$  0.25}   &  71.29 {\tiny  $\pm$  0.27}\\
 $\eps=0.075$      &  74.70 {\tiny  $\pm$  0.26} &  75.22 {\tiny  $\pm$  0.21} &  \textbf{75.62} {\tiny  $\pm$  0.15}   &  75.13 {\tiny  $\pm$  0.19}\\
 $\eps=0.10$      &  75.56 {\tiny  $\pm$  0.18} &  76.91 {\tiny  $\pm$  0.17} &  \textbf{77.65} {\tiny  $\pm$  0.20}     &  77.05 {\tiny  $\pm$  0.09}\\
 $\eps=0.20$      &  79.44 {\tiny  $\pm$  0.07} &  80.23 {\tiny  $\pm$  0.12} &  \textbf{80.76} {\tiny  $\pm$  0.08}     &  80.37 {\tiny  $\pm$  0.13}\\
 $\eps=0.30$      &  80.19 {\tiny  $\pm$  0.17} &  81.70 {\tiny  $\pm$  0.08} &  82.15 {\tiny  $\pm$  0.15}              &  \textbf{82.19} {\tiny  $\pm$  0.10}\\
 $\eps=0.40$      &  81.91 {\tiny  $\pm$  0.10} &  82.86 {\tiny  $\pm$  0.05} &  \textbf{82.82} {\tiny  $\pm$  0.13}              &  82.80 {\tiny  $\pm$  0.09}\\
 $\eps=0.60$      &  82.56 {\tiny  $\pm$  0.36} &  82.91 {\tiny  $\pm$  0.20} &  83.44 {\tiny  $\pm$  0.15}              &  \textbf{83.46} {\tiny  $\pm$  0.04}\\
 $\eps=0.80$      &  83.40 {\tiny  $\pm$  0.06} &  83.67 {\tiny  $\pm$  0.22} &  84.10 {\tiny  $\pm$  0.07}              &  \textbf{84.44} {\tiny  $\pm$  0.07}\\
 $\eps=2.00$      &  84.34 {\tiny  $\pm$  0.08} &  84.80 {\tiny  $\pm$  0.18} &  85.18 {\tiny  $\pm$  0.08}              &  \textbf{85.97} {\tiny  $\pm$  0.04}\\
 $\eps=4.00$      &  84.82 {\tiny  $\pm$  0.04} &  85.11 {\tiny  $\pm$  0.08} &  85.40 {\tiny  $\pm$  0.08}              &  \textbf{86.63} {\tiny  $\pm$  0.05}\\
 $\eps=8.00$      &  85.13 {\tiny  $\pm$  0.05} &  85.37 {\tiny  $\pm$ 0.04} & 85.62 {\tiny  $\pm$ 0.04}                  &  \textbf{87.05} {\tiny  $\pm$  0.03}\\
 $\eps=\infty$      &  \m{85.19 {\tiny  $\pm$  0.3}} &  \m{85.42 {\tiny  $\pm$  0.02}} &  \m{85.80 {\tiny  $\pm$  0.03}}          &  \m{\textbf{87.85} {\tiny  $\pm$  0.02}}\\
    \bottomrule
    \end{tabular}
\end{subtable}
\vspace{-1em}
\end{table}

\begin{table*}[t]
\centering
\caption{\small Model utility for applying norm-based \top and dimension-based \top in DP SGD training on MNIST dataset. 
$\eps$ is the privacy budget and $k$ is the top-$k$ parameter in \top. 
In all cases, $\delta=10^{-5}.$ 
}\label{tab:dpsgd-ablation}

\begin{subtable}[c]{\linewidth}
	\centering
    \begin{tabular}{l|ccc|ccc|c}
    \toprule
        & \multicolumn{3}{c|}{\small \textbf{\top} (norm-based)} & \multicolumn{3}{c|}{\small \textbf{\top} (dim-based)} & \multirow{2}{*}{\small \textbf{\gmdp}}  \\
    & \multicolumn{1}{c}{\small $k=0.6$}  & \multicolumn{1}{c}{\small $k=0.7$} & \multicolumn{1}{c|}{\small $k=0.8$} & \multicolumn{1}{c}{\small $k=0.2$}  & \multicolumn{1}{c}{\small $k=0.4$} & \multicolumn{1}{c|}{\small $k=0.6$} &   \\\midrule
    $\eps=0.05$      &  73.87 {\tiny  $\pm$  4.77} &  75.81 {\tiny  $\pm$  3.59} &  77.15 {\tiny  $\pm$  2.89}             
                    & 74.25 {\tiny  $\pm$  3.25}  & \textbf{78.45} {\tiny  $\pm$  2.89}  &  73.55 {\tiny  $\pm$  1.97}           
                    &  78.40 {\tiny  $\pm$  4.00}\\
    $\eps=0.10$     &  85.67 {\tiny  $\pm$  1.48}  &  86.12 {\tiny  $\pm$  1.39} &  85.88 {\tiny  $\pm$  1.97}    
                    & 83.65 {\tiny $\pm$ 0.73} &  85.59 {\tiny $\pm$ 1.39} & \textbf{86.41} {\tiny $\pm$ 0.70}            
                    &  84.55 {\tiny  $\pm$  0.98}\\
    $\eps=0.20$     &  89.84 {\tiny  $\pm$  0.64}  &  90.25 {\tiny  $\pm$  0.40} &  90.80 {\tiny  $\pm$  1.04}             
                    & 89.10 {\tiny  $\pm$  1.22}  &  90.48 {\tiny  $\pm$  0.73} &  89.30 {\tiny  $\pm$  0.74}             
                    &  \textbf{91.17} {\tiny  $\pm$  0.37}
    \\
    \bottomrule
    \end{tabular}
\end{subtable}
\end{table*}

\label{sec:exp-sgd-results} 
We compare \top (with $k\in\{0.7, 0.8, 1.0\}$) with two baselines (\gmdp and DPDL) on DP SGD training under a wide variety of privacy constraints, as shown in~\Cref{fig:dpsgd}. 
For clarity of presentation, we leave the complete set of results in~\Cref{tab:dpsgd}.
For MNIST, we mainly evaluate small privacy budgets ($\eps\leq 1.0$), since the performance gap of private and non-private models on MNIST is negligible for large $\varepsilon$.
For CIFAR-10, specifically, we examine the regions of both small $\eps$ and large $\eps$ respectively in a more comprehensive manner.

We first note that \gmdp is a special case of our \top when $k=1.0$. In this case, the complete gradient is preserved after the Top-K step, and therefore the performance of \top is equivalent to \gmdp.
Still, we provide both results as a sanity check in~\Cref{fig:dpsgd}. We observe that
the results of \top ($k=1.0$) and \gmdp are indeed close in almost all scenarios regardless of the randomness of the algorithm (the green and red lines are generally overlapped).

In~\Cref{fig:dpsgd} (a) for MNIST, different curves are intertwined, indicating that choosing different $k$ values can only influence the model performance by a little margin.
Thus, it does not hurt to adopt \top with different $k$. The main reason is that, on MNIST, the model utility gap before and after adding DP noise is small, which does not provide space for smaller gradient compression ratios to improve the performance.

Moreover, for CIFAR-10 \m{whose dimensionality is around 4 times larger than MNIST}, we observe consistent performance improvements under the limited privacy budget $\eps$ in~\Cref{fig:dpsgd} (b), which is also aligned with our observation in \top for generative models, where our methods demonstrate a large margin over baselines under limited privacy budgets. Specifically, In~\Cref{tab:dpsgd} (b) under $\varepsilon=0.025$, we observe that \top  with $k=0.8$ can achieve the model accuracy of $45.87\%$, which is more than $4\%$ higher than the baseline. This again verifies our theoretical analysis that \top can help save the privacy budget consumption by compressing the gradient, and therefore substantially help the model convergence and improve the utility of the model. 

With larger privacy budgets, \Cref{fig:dpsgd} (c) shows that \top with very small $k$ tends to have worse performance. It indicates that given small $k$, the bias introduced by top-$k$ compression outweighs the bias introduced by low differential privacy noise. Without DP noise ($\varepsilon=\infty$), we observe that \top with $k < 1$ has slightly worse  performance than the \gmdp baseline, which is because without DP noise there is no longer the benefits of lower DP noise brought by top-$k$, while the bias introduced by gradient compression starts to hurt the performance moderately.

We further point out that DPDL has the worst performance among all as shown in~\Cref{fig:dpsgd}. This phenomenon is well understood given that the privacy analysis is looser in Abadi et al.~\citep{Abadi2016DeepLW} compared with the privacy analysis based on R\'enyi Differential Privacy~\citep{mironov2019r} adopted in both our \top and \gmdp here.

In addition, we empirically examine the hyper-parameters and the impact of gradient compression and noise injection in \top on MNIST.  We omit the detailed results in~\Cref{sec:appen-top}. We observe that the bias induced by the DP noise is indeed larger than the gradient compression, which confirms our theoretical analysis of the convergence for \top.

\input{tables/tab_control}

\begin{table*}[t]
\vspace{-1em}
\centering
\caption{\small Optimal hyper-parameters for \top and GM-DP baseline on MNIST and CIFAR-10 with different privacy parameter $\eps$ and \top parameter $k$. In all cases, $\delta=10^{-5}.$ 
}\label{tab:dpsgd-search}

\begin{subtable}[]{0.31\linewidth}
	\centering
	\vspace{0.5em}
	\caption{\small MNIST ($\varepsilon=0.05$)}
	\resizebox{0.9\columnwidth}{!}{%
    \begin{tabular}{l|cccc}
    \toprule
    & \multicolumn{3}{c}{\small \textbf{\top}} & \multirow{2}{*}{\small \textbf{\gmdp}}  \\
			& \multicolumn{1}{c}{\small $k=0.6$}  & \multicolumn{1}{c}{\small $k=0.7$} & \multicolumn{1}{c}{\small $k=0.8$}     &   \\\midrule
  $c$           &   10.0    &    5.0  &      5.0      &  3.0 \\
 $\sigma$       &  5.8    &     6.6 &        7.4       &  8.2  \\
 batch size     &  128    &     128      &  128      & 128 \\
 learning rate  &  0.01    &     0.01 &      0.01   & 0.01 \\
    \bottomrule
    \end{tabular}
	}
\end{subtable}
\begin{subtable}[]{0.31\linewidth}
	\centering
	\vspace{0.5em}
	\caption{\small MNIST ($\varepsilon=0.2$)}
	\resizebox{0.9\columnwidth}{!}{%
    \begin{tabular}{l|cccc}
    \toprule
    & \multicolumn{3}{c}{\small \textbf{\top}} & \multirow{2}{*}{\small \textbf{\gmdp}}  \\
			& \multicolumn{1}{c}{\small $k=0.6$}  & \multicolumn{1}{c}{\small $k=0.7$} & \multicolumn{1}{c}{\small $k=0.8$}     &   \\\midrule
  $c$           &   12.0    &    10.0      &   10.0      &  10.0 \\
 $\sigma$       &  3.6    &     3.6      &    3.6       &  3.4  \\
 batch size     &  512    &    512     &   512     & 512 \\
 learning rate  &  0.1    &     0.1      &   0.1      & 0.08 \\
    \bottomrule
    \end{tabular}
	}
\end{subtable}
\begin{subtable}[]{0.31\linewidth}
	\centering
	\vspace{0.5em}
	\caption{\small MNIST ($\varepsilon=1.0$)}
	\resizebox{0.9\columnwidth}{!}{%
    \begin{tabular}{l|cccc}
    \toprule
    & \multicolumn{3}{c}{\small \textbf{\top}} & \multirow{2}{*}{\small \textbf{\gmdp}}  \\
			& \multicolumn{1}{c}{\small $k=0.6$}  & \multicolumn{1}{c}{\small $k=0.7$} & \multicolumn{1}{c}{\small $k=0.8$}     &   \\\midrule
 $c$           &   16.0    &    12.0 &         10.0      &  14.0 \\
 $\sigma$      &   1.6    &     1.6 &        1.6       &  1.8  \\
 batch size    &   512    &     512    &   512       & 512 \\
 learning rate &   0.08    &     0.1 &      0.1   & 0.04 \\
    \bottomrule
    \end{tabular}
	}
\end{subtable}
\begin{subtable}[]{0.31\linewidth}
	\centering
	\vspace{0.5em}
	\caption{\small CIFAR-10 ($\varepsilon=0.025$)}
	\resizebox{0.9\columnwidth}{!}{%
    \begin{tabular}{l|cccc}
    \toprule
    & \multicolumn{3}{c}{\small \textbf{\top}} & \multirow{2}{*}{\small \textbf{\gmdp}}  \\
			& \multicolumn{1}{c}{\small $k=0.6$}  & \multicolumn{1}{c}{\small $k=0.7$} & \multicolumn{1}{c}{\small $k=0.8$}     &   \\\midrule
 $c$           &   1    &    1.5 &         0.8      &  0.3 \\
 $\sigma$      &   6    &     7 &        6       &  7  \\
 batch size    &   24    &     32    &   24       & 32 \\
 learning rate &   0.0008    &     0.001 &      0.001   & 0.002 \\
    \bottomrule
    \end{tabular}
	}
\end{subtable}
\begin{subtable}[]{0.31\linewidth}
	\centering
	\vspace{0.5em}
	\caption{\small CIFAR-10 ($\varepsilon=0.4$)}
	\resizebox{0.9\columnwidth}{!}{%
    \begin{tabular}{l|cccc}
    \toprule
    & \multicolumn{3}{c}{\small \textbf{\top}} & \multirow{2}{*}{\small \textbf{\gmdp}}  \\
			& \multicolumn{1}{c}{\small $k=0.6$}  & \multicolumn{1}{c}{\small $k=0.7$} & \multicolumn{1}{c}{\small $k=0.8$}     &   \\\midrule
  $c$           &   1    &    1  &      1      &  1.5 \\
 $\sigma$       &  2.5    &     2.5 &        2.5       &  3  \\
 batch size     &  96    &     96      &  96      & 96 \\
 learning rate  &  0.005    &     0.005 &      0.005   & 0.005 \\
    \bottomrule
    \end{tabular}
	}
\end{subtable}
\begin{subtable}[]{0.31\linewidth}
	\centering
	\vspace{0.5em}
	\caption{\small CIFAR-10 ($\varepsilon=8$)}
	\resizebox{0.9\columnwidth}{!}{%
    \begin{tabular}{l|cccc}
    \toprule
    & \multicolumn{3}{c}{\small \textbf{\top}} & \multirow{2}{*}{\small \textbf{\gmdp}}  \\
			& \multicolumn{1}{c}{\small $k=0.6$}  & \multicolumn{1}{c}{\small $k=0.7$} & \multicolumn{1}{c}{\small $k=0.8$}     &   \\\midrule
  $c$           &   0.5    &    0.5      &   0.5      &  2.5 \\
 $\sigma$       &  2.5    &     2.5      &    3       &  2.5  \\
 batch size     &  2048    &    2048     &   2048     & 2048 \\
 learning rate  &  0.2    &     0.2      &   0.2      & 0.04 \\
    \bottomrule
    \end{tabular}
	}
\end{subtable}
\vspace{-1em}
\end{table*}

\subsection{Ablation Studies on Hyper-parameters} 

DP-SGD algorithms (GM-DP and DPDL) contains several key parameters: the noise multiplier $\sigma$ of the injected Gaussian noise, gradient clipping constant $c$, batch size that will affect the sampling rate $q$, and learning rate. \top adds another important parameter $k$ for \texttt{NormTopK} on top of the GM-DP framework. To search for the optimal hyper-parameters, we conduct comprehensive grid search. We list the optimal hyper-parameters under several different privacy budgets $\varepsilon$ in~\Cref{tab:dpsgd-search} for MNIST and CIFAR-10. 

\subsection{Tradeoff between Gradient Compression and Noise Injection}\quad
\label{sec:appen-tradeoff}
\vspace{-0.3em}

In essence, \top differs from 
the standard DP SGD training scheme and moment accountant method 
adopted in \gmdp~\citep{mcmahan2019tfprivacy}  mainly in the introduction of the gradient compression parameter $k$.
On the one hand, using a smaller $k$ to compress the gradients leads to more biased results in the returned gradients, which can cause performance degradation. On the other hand, the gradient norm becomes smaller after \top based training which enables the introduction of less noise to achieve the same level of privacy guarantee, and therefore less distortion to the prediction.
Noticing the tradeoff, we ask, \textit{is there  a sweet spot where the performance increase caused by the injection of less noise surpasses the model utility degradation induced by the bias in gradient compression?}

To this end, we design a set of control experiments to analyze the impact of the two factors: 
1) compression parameter of gradients and 2) amount of injected DP noise.
We provide the setup of the control experiments in~\Cref{tab:control-setup}. 
Concretely, we investigate 3 levels of noise injection corresponding to the requirement of 3 algorithms (non-private SGD~\citep{Kiefer1952StochasticEO}, \gmdp~\citep{mcmahan2019tfprivacy}, and our \top), and 2 scenarios of gradient compression (no compression and our \texttt{NormTopK} compression).
Note that only \gmdp and \top satisfy the intended privacy requirements.
We do not report the results for the combination of no compression and reduced noise, since it is blatantly non-private and does not offer additional insights. Rather, we investigate the combination of \texttt{NormTopK} and full noise which we title TopK-\gmdp, in a hope to build the bridge between \gmdp and \top.
We additionally examine TopK-SGD (the combination of \texttt{NormTopK} and no noise) to get the sense of an upper-bound of the performance after gradient compression.
For the purpose of controlling variables, we control the clipping norm $C$ and noise scale $\sigma$ to be the same for all the scenarios, and train the non-private algorithms using the same number of iterations as the private ones, which is determined by the corresponding privacy budget.

The results of the control experiments on different privacy budgets are provided in~\Cref{tab:control-res} (b). We summarize our observations as follows.
First, along each row or each column of~\Cref{tab:control-res} (b), the performance of the training schemes will experience an increase. Naturally, the non-private SGD training will give the best performance of all (despite norm clipping). 
This means that less noise and no compression will generally yield better results.
Second, noise injection has a larger impact on the performance than gradient compression in both cases of small and big $\eps$. 
Third, the reduction in the scale of injected noise will compensate the performance decrease caused by gradient compression, therefore resulting in a similar or slightly better performance of \top when compared with \gmdp.
Given these observations, we conclude that the impact of gradient compression is negligible compared with noise injection, and that it is beneficial to exploit gradient compression to trade for the reduction in injected noise to achieve a potentially better performance.

\section{Visualization of Image Quality}

\begin{figure}[t]
    \centering
    \includegraphics[width=\linewidth]{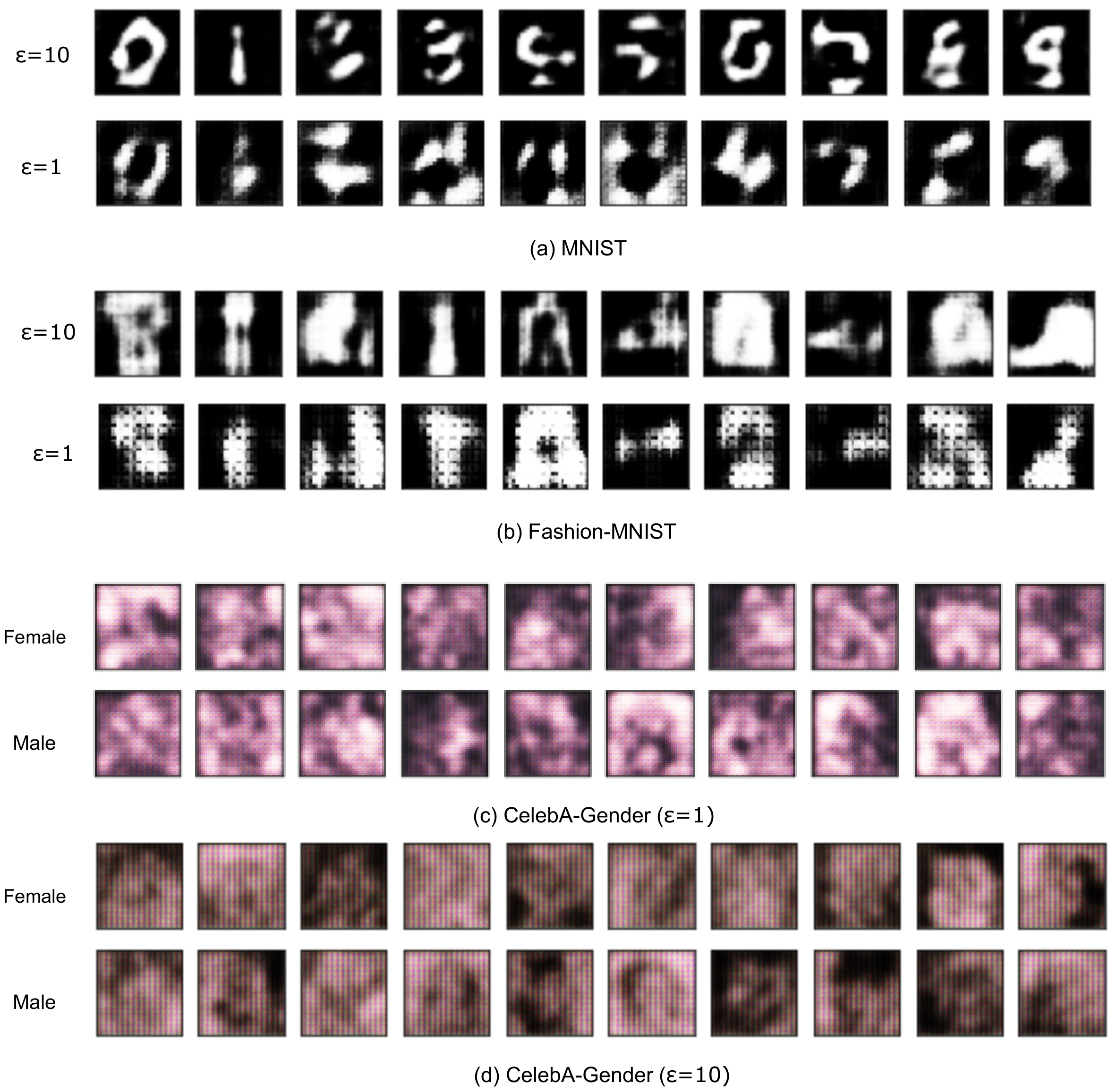}
    \caption{Visualization of generated images from DataLens}
    \label{fig:vis}
\end{figure}

We visualize the private synthetic images for MNIST, Fashion-MNIST and CelebA, as shown in~\Cref{fig:vis}. As expected the image has a lot of noise, since our goal is to generate data which can protect privacy and ensure high data utility in terms of training high performance models.  It is interesting to see that these generated images are enough to train useful models, which lead to interesting future direction on what ML models actually learn from data.

\section{Experimental Details}

\begin{small}
\begin{algorithm}[t]
\caption{\textbf{- Gradient Compression  via k-level Stochastic  Gradient (\texttt{StoKlevelGrad}).} This algorithm takes in a gradient vector of a teacher model $\mathbf{g}^{(i)}$ and returns the compressed gradient vector $\mathbf{\tilde{g}}^{(i)}$.}
\label{algo:klevel}
\begin{algorithmic}[1]
    \State {\bfseries Input:}  Gradient vector $\mathbf{g}^{(i)}$, gradient clipping constant $c$, top-$k$
  
    \State $\mathbf{g}^{(i)}_j = \min(\max(\mathbf{g}^{(i)}_j, -c), c)$ for each dimension $j$ in ${\mathbf{g}}^{(i)}$
    \Statex \hfill\Comment{Clip each dimension of $\mathbf{g}^{(i)}$ so that $-c \le \mathbf{g}_j^{(i)} \le c.$} 
    \State $\mathbf{g}^{(i)} = R \times \mathbf{g}^{(i)}$ \hfill \Comment{Random Rotation}
    \State $\hat{\mathbf{g}}^{(i)} \leftarrow \mathbf{g}^{(i)} /  \left\|\mathbf{g}^{(i)} \right\|_{\infty}$ \quad
    \hfill\Comment{gradient normalization to (-1, 1)} 
    \State $\mathbf{\tilde{g}}^{(i)} \leftarrow \mathbf{0}$
    \State let $b[r] := -k / 2 + 2r$ for every $r \in [0, k)$
    \State let $m[r] := -c + \frac{2rc}{k-1}$ for every $r \in [0, k)$
    \hfill\Comment{initialization of the compressed sparse gradient vector}
    \For{each index $j$, and $b[r] \le \mathbf{g}^{(i)}_j \le b[r+1]$ } 
        \State ${\tilde{g}}_j^{(i)} =\left\{
                \begin{array}{ll}
                  b[r+1], & \textit{with probability } \frac{\mathbf{g}^{(i)}_j - m[r]}{m[r+1] - m[r]}\\
                  b[r], & o.w. \\
                \end{array}
              \right.$
    \EndFor
    \State {\bfseries Return:} $\mathbf{\tilde{g}}^{(i)}$
\end{algorithmic}
\end{algorithm}
\end{small}


\begin{small}
\begin{algorithm}[t]
\caption{\textbf{- Differentially Private Gradient Compression and Aggregation (\dtwop for \sysname).} This algorithm takes gradients of teacher models and returns the compressed and aggregated differentially private gradient vector.}
\label{algo:d2p}
\begin{algorithmic}[1]
    \State {\bfseries Input:} Teacher number $N$, gradient vectors of teacher models $\mathcal{G}=\{\mathbf{g}^{(1)}, \dots, \mathbf{g}^{(N)}\}$, gradient clipping constant $c$, top-$k$, noise parameters $\sigma$, voting threshold $\beta$
    \State\Comment{Phase I: Gradient Compression}
    \For{each teacher's gradient $\mathbf{g}^{(i)}$}
        \State $\mathbf{\tilde{g}}^{(i)} \leftarrow \texttt{StoKlevelGrad}(\mathbf{g}^{(i)}, c, k) $
    \EndFor
    \State\Comment{Phase II: Differential Private Gradient Aggregation}
    \State $\mathbf{\tilde{g}}^* \leftarrow \sum_{i=1}^N {\mathbf{\tilde{g}}^{(i)}}  + \mathcal{N}(0, \sigma^2)$
    \State\Comment{Phase III: Gradient Thresholding (Post-Processing)}
    \For{each dimension ${\tilde{g}}_j^*$ of $\mathbf{\tilde{g}}^*$}
        \State ${{\bar{g}}}_j =\left\{
                \begin{array}{ll}
                  1, &  \text{if} \quad  {\tilde{g}}_j^* \ge \beta N ;\\
                  -1, & \text{if} \quad  {\tilde{g}}_j^* \le -\beta N ;\\
                  0, & \text{otherwise.}
                \end{array}
              \right.$
    \EndFor
    \State {\bfseries Return:} $\mathbf{\bar{g}}$
\end{algorithmic}
\end{algorithm}
\end{small}

\begin{small}
\begin{algorithm}[t]
\caption{\textbf{- Differentially Private Gradient Compression and Aggregation (\fetch for \sysname).} This algorithm takes gradients of teacher models and returns the compressed and aggregated differentially private gradient vector.}
\label{algo:fetch}
\begin{algorithmic}[1]
    \State {\bfseries Input:} Teacher number $N$, gradient vectors of teacher models $\mathcal{G}=\{\mathbf{g}^{(1)}, \dots, \mathbf{g}^{(N)}\}$, gradient clipping constant $c$, top-$k$, noise parameters $\sigma$, voting threshold $\beta$
    \State S = CountSketchAggregator()
    \State\Comment{Phase I: Gradient Compression}
    \For{each teacher's gradient $\mathbf{g}^{(i)}$}
        \State $\mathbf{\tilde{g}}^{(i)} \leftarrow \texttt{TopkStoSignGrad}(\mathbf{g}^{(i)}, c, k) $
        \State S += Sketch($\mathbf{\tilde{g}}^{(i)}$)
    \EndFor
    \State\Comment{Phase II: Differential Private Gradient Aggregation}
    \State $\mathbf{\tilde{g}}^* \leftarrow \text{top-}k(\text{unSketch}(S))  + \mathcal{N}(0, \sigma^2)$
    \State {\bfseries Return:} $\mathbf{\tilde{g}}^* $
\end{algorithmic}
\end{algorithm}
\end{small}

\subsection{Visual Quality Evaluation}

We evaluate both Inception Score and Frechet Inception Distance for \sysname and baselines over four datasets. We present the evaluation results in Table \ref{tab:full_image}.

For Inception Score, in our experiments, we follow GS-WGAN and use the implementation\footnote{\url{https://github.com/ChunyuanLI/MNIST_Inception_Score}} for Inception Score calculation with pretrained classifiers trained on real datasets (with test accuracy equal to $99\%, 93\%, 97\%$ on MNIST, Fashion-MNIST, and CelebA-Gender).  

For FID, we observe it is not necessarily consistent with Inception Score (e.g., for MNIST $\varepsilon=1$, \sysname has better IS than G-PATE, but worse FID than G-PATE), which we think the reason is because FID is evaluated based on models trained with ImageNet which may not be suitable for evaluating datasets such as MNIST. In our experiments, we follow GS-WGAN and use the implementation\footnote{\url{https://github.com/google/compare_gan}} for FID calculation.

\subsection{Adapting Other Gradient Compression Algorithms to \sysname} \label{sec:app-compression}

In this section, we illustrate how we adapt \dtwop and \fetch to \sysname framework.

For \dtwop, we replace our Algorithm 2 (\texttt{TopkStoSignGrad}) that uses stochastic sign compression with their method, which essentially uses k-level gradient quantization and random rotation for gradient pre-processing. The detailed algorithm is shown in Algorithm \ref{algo:d2p} and Algorithm \ref{algo:klevel}.

For \fetch, we uses the same stochastic sign compression as we leverage sign signal as teacher voting in PATE framework. During aggregation, we use Count Sketch data structure, and use top-$k$ and unsketch operation to retrieve the aggregated gradient. The detailed algorithm is shown in Algorithm \ref{algo:fetch}.

\clearpage
\onecolumn
\begin{table*}[t!]
\centering
\caption{\small \textbf{Quality evaluation of images generated by different differentially private data generative models on Image Datasets:} 
Inception Score (IS) and Frechet Inception Distance (FID) are calculated to measure the visual quality of the generated data  under different $\eps$ ($\delta=10^{-5}$). %
}
\begin{subtable}[h]{\linewidth}
\centering
\caption{$\varepsilon=1$}
\m{
\begin{tabular}{l|c|c|ccccc}
\toprule
\textbf{Dataset} & Metrics &\textbf{Real data}  &\textbf{DP-GAN} & \textbf{PATE-GAN}  &\textbf{G-PATE} & \textbf{GS-WGAN} &\textbf{\name{}}  \\ \midrule
\multirow{2}{*}{\textbf{MNIST}} & IS $\uparrow$ & 9.86 &  1.00 & 1.19  & 3.60 & 1.00 & \textbf{{4.37}}  \\ 
& FID $\downarrow$ & 1.04 & 470.20 & {231.54} & \textbf{153.38} &  489.75  & 186.06  \\ \midrule
\multirow{2}{*}{\textbf{Fashion-MNIST}} & IS $\uparrow$ & 9.01 &  1.03
& 1.69 & 3.41 & 1.00 & \textbf{{3.93}}  \\ 
& FID $\downarrow$ & 1.54 & 472.03 & 253.19 &  214.78  &  587.31  & \textbf{194.98} \\ \midrule
\multirow{2}{*}{\textbf{CelebA}} & IS $\uparrow$ & 1.88 &  1.00
& 1.15 & 1.11 & 1.00 & \textbf{{1.18}}  \\ 
& FID $\downarrow$ & 2.38 &  485.92 & 434.47 & 302.45  &  437.33  & \textbf{297.73}  \\ 
\bottomrule
\end{tabular}%
}
\end{subtable}

\vspace{0.8em}
\hfill

\begin{subtable}[h]{\linewidth}
\centering
\caption{$\varepsilon=10$}
\m{
\begin{tabular}{l|c|c|ccccc}
\toprule
\textbf{Dataset} & Metrics &\textbf{Real data}  &\textbf{DP-GAN} & \textbf{PATE-GAN}  &\textbf{G-PATE} & \textbf{GS-WGAN} &\textbf{\name{}}  \\ \midrule
\multirow{2}{*}{\textbf{MNIST}} & IS $\uparrow$ & 9.86 &  1.00
& 1.46 & 5.16 &  \textbf{8.59} & 5.78  \\ 
& FID $\downarrow$ & 1.04 & 304.86 & {253.55} & 150.62 &  \textbf{58.77}  & 173.50  \\ \midrule
\multirow{2}{*}{\textbf{Fashion-MNIST}} & IS $\uparrow$ & 9.01 &  1.05
& 2.35 & 4.33 & \textbf{5.87}  & 4.58  \\ 
& FID $\downarrow$ & 1.54 & 433.38 & 229.25 & 171.90 &  \textbf{135.47}  & 167.68  \\ \midrule
\multirow{2}{*}{\textbf{CelebA}} & IS $\uparrow$ &  1.88 &  1.00
& 1.16 & 1.12 & 1.00 & \textbf{{1.42}}  \\
& FID $\downarrow$ & 2.38 &  485.41  & 424.60 &  323.95  &  432.58  & \textbf{320.84} \\
\bottomrule
\end{tabular}%
}
\end{subtable}

\label{tab:full_image}
\vspace{-0.5em}
\end{table*}

\section{Proofs}\label{sec:app_proofs}
We now prove~\Cref{theorem:data_dependent_prob} and~\Cref{thmConvergenceAnalysis}.

\paragraph{Proof of~\Cref{theorem:data_dependent_prob}.}
We have
\begin{equation*}
\begin{aligned}
    & \Pr[\mathcal{M}(\mathcal{\tilde{G}}, T, \beta) \neq \mathbf{\bar{g}}^*] \\
    = & 1 - \Pr[\mathcal{M}(\mathcal{\tilde{G}}, T, \beta) = \mathbf{\bar{g}}^*] \\
    = & 1 - \prod_{\{j \mid \bar{g}^*_j=1\}}\Pr[f_j+n_j \geq \beta T]
    \prod_{\{j \mid \bar{g}^*_j=-1\}}\Pr[f_j+n_j \leq -\beta T]
    \prod_{\{j \mid \bar{g}^*_j=0\}}\Pr[-\beta T < f_j+n_j < \beta T]\\
    = & 1 - 
    \prod_{\{j \mid \bar{g}^*_j=1\}}\Pr[n_j \geq \beta T - f_j]
    \prod_{\{j \mid \bar{g}^*_j=-1\}}\Pr[n_j \leq -\beta T - f_j]
    \prod_{\{j \mid \bar{g}^*_j=0\}} \Pr[-\beta T - f_j < n_j < \beta T - f_j] \\
    = & 1 - 
    \prod_{\{j \mid \bar{g}^*_j=1\}}\left(1-\Phi\left(\frac{\beta T-f_j}{\sigma}\right)\right)
    \prod_{\{j \mid \bar{g}^*_j=-1\}}\Phi\left(\frac{\beta T - f_j}{\sigma}\right)
    \prod_{\{j \mid \bar{g}^*_j=0\}}\erf\left(\frac{\beta T - f_j}{\sqrt{2}\sigma}\right),
\end{aligned}
\end{equation*}
where the last equality holds because $n_j$ follows the normal distribution with mean $0$ and variance $\sigma^2$, concluding the proof.

\paragraph{Proof of~\Cref{thmConvergenceAnalysis}.}

We begin by fixing $t\in [T]$. The assumption that $f$ has $L$-Lipschitz gradient, \ie, $\|\nabla f(x) - \nabla f(y)\| \leq L \|x-y\|$, implies, through a well-known argument, that 
\[
f(x_{t+1}) - f(x_t) \leq \inn{\nabla f(x_t)}{x_{t+1} - x_t} + \frac{L}{2} \|x_{t+1} - x_t\|^2,
\] 
Recall that 
$x_{t+1} - x_t  = -  \frac{\gamma}{N} \sum_{n\in[N]} \left(Q ({\kk\texttt{clip}}(\texttt{top-k}(F_{n}' (x_t)),c), \xi_t) + \mathcal{N} (0, Ak)\right).$
Taking the expectation over the quantization and the insertion of data-privacy noise yields
\begin{align*}
	\E_\cN \E_{\xi_t} f(x_{t+1}) - f(x_t) &\leq -\frac{\gamma}{N}\sum_{n\in[N]}\underbrace{\inn{\nabla f(x_t)}{\E_{\xi_t} \left[ Q({\kk\texttt{clip}}(\texttt{top-k} (F_{n}' (x_t),c), \xi_t)\right]}}_{I_n(x_t)} - \underbrace{\frac{\gamma}{N}\sum_{n\in[N]}\inn{\nabla f(x_t)}{\E_\cN \left[ \cN(0,Ak) \right]}}_{ = 0}\\ 
	&\hspace{20mm} + \frac{L\gamma^2}{N}\sum_{n\in [N]} \underbrace{\E_{\xi_t}\left\| Q({\kk\texttt{clip}}(\texttt{top-k}(F_n'(x_t)),c),\xi_t) \right\|^2}_{J_n(x_t)} + \underbrace{\frac{L\gamma^2}{N} \sum_{n\in [N]}\E_\cN \| \cN(0,Ak)\|^2}_{=L\gamma^2 Ak},
\end{align*}
where we used the Cauchy-Schwarz inequality $(a_1 + \ldots + a_n)^2 \leq n (a_{1}^2 + \ldots + a_{n}^2)$ for $n=2N$.

For $I_n(x_t)$ note that 
	\begin{align*}
	-\frac{\gamma}{N} \sum_{n\in [N]} I_n (x_t) &= -\frac{\gamma}{N}\inn{\nabla f(x_t)}{\E_{\xi_t} \left[ Q({\kk\texttt{clip}}(\texttt{top-k} (F_{n}' (x_t)),c), \xi_t)\right]} \\
	\left\{ \E_\xi \left[Q(x,\xi)\right] = x \right\} \hspace{10mm}&= -\frac{\gamma}{N} \sum_{n\in [N]} \inn{\nabla f(x_t)}{{\kk\texttt{clip}}(\texttt{top-k}(F_{n}' (x_t)),c)}\\
	&{\kk= -\frac{\gamma}{N} \sum_{n\in [N]} \underbrace{\inn{\nabla f(x_t)}{\texttt{clip}(F_{n}' (x_t),c)}}_{I_{n}^{(1)}} +\frac{\gamma}{N} \sum_{n\in [N]} \underbrace{\inn{\nabla f(x_t)}{\texttt{clip}(F_{n}' (x_t)),c)-\texttt{clip}(\texttt{top-k}(F_{n}' (x_t)),c)}}_{I_{n}^{(2)}}}
    \end{align*}
    
 {\kk
 \textbf{Claim 1.} For $\alpha=\frac{1}{d+2}$ and under the assumptions from Theorem~\ref{thmConvergenceAnalysis} one has
 \begin{equation}\label{eqn:mainclaim}
     -\frac{\gamma}{N}\sum_{n\in[N]}I_{n}^{(1)} \leq \gamma \max\{-\alpha \|\nabla f(x_t) \|^2 + \|\sigma\|^2 + \|\sigma\|M, \hspace{2mm}-\alpha c \|\nabla f(x_t)\|_1 + 2c\|\sigma\|_1\}.
 \end{equation}
 
\textit{Proof of Claim 1.} 	For the ease of notation, let $x=x_t$ and $g_{n}(x)=F_{n}' (x)$. First note that, per coordinate $i \in [d]$,
	\[
	\texttt{clip}(g_{n}(x)_i,c) = c \cdot  \texttt{sign} (g_{n}(x)_i) \cdot \mathbf{1} \{|g_{n}(x)_i|\geq c\} + g_{n}(x)_i \cdot \mathbf{1} \{|g_{n}(x)_i|< c\}.
	\]
	
	The main idea is to prove that one of these yields the main term, which would correspond to $-\gamma \|\nabla f(x) \|^2$ for the usual gradient descent, and $-c\gamma \| \nabla f(x)\|_1$ for the signed gradient descent. With that in mind, let us for each $i\in[d]$ define $A_i = \{ n \in [N]\colon |g_{n}(x)_i| \geq c\}$ and $B_i = \{ n \in [N]\colon |g_{n}(x)_i| < c\}$, with $A_i \cap B_i = \emptyset$ and $A_i \cup B_i = [N]$, for all $i\in [d]$. Then 
	\[
	-\frac{\gamma}{N}\sum_{n\in[N]}I_{n}^{(1)} =-\frac{\gamma c}{N}\sum_{i\in[d]}\sum_{n\in A_i} \nabla f(x)_i \cdot \texttt{sign}(g_n(x)_i) -\frac{\gamma}{N}\sum_{i\in[d]}\sum_{n\in B_i} \nabla f(x)_i \cdot g_n(x)_i.
	\]
	To explore the above mentioned dichotomy, we now rewrite the quantity we are trying to estimate in two ways:
	\begin{align*}
-\frac{\gamma}{N}\sum_{n\in[N]}I_{n}^{(1)}  &= -\gamma \| \nabla f(x)\|^2 + \underbrace{\frac{\gamma}{N}\sum_{i\in[d]} \sum_{n \in A_i} |\nabla f(x)_i|^2}_{\texttt{err}(GD)}+\underbrace{\frac{\gamma}{N}\sum_{i\in[d]}\sum_{n\in A_i} \nabla f(x)_i \left( g_{n}(x)_i - \nabla f(x)_i -  c\cdot \texttt{sign} (g_{n}(x)_i) \right)}_{\texttt{err}(GD_2)},\\
-\frac{\gamma}{N}\sum_{n\in[N]}I_{n}^{(1)}  &= -c\gamma \| \nabla f(x)\|_1 + \underbrace{\frac{\gamma}{N}\sum_{i\in[d]} \sum_{n \in B_i} \nabla f(x)_i \left(c\cdot \texttt{sign} (g_{n}(x)_i) - g_{n}(x)_i\right)}_{\texttt{err}(signGD)}+  \underbrace{\frac{\gamma c}{N} \sum_{i \in [d] } \sum_{ n\in [N] } \nabla f(x)_i \left( \texttt{sign} (\nabla f(x)_i) - \texttt{sign} (g_{n}(x)_i) \right)}_{\texttt{err}(signGD_2)}.
	\end{align*}
	
	We start by bounding $\texttt{err}(GD_2)$ and $\texttt{err}(singGD_2)$. For $\texttt{err}(GD_2)$ we have 
	\begin{align*}
	    \texttt{err}(GD_2) &\leq \frac{\gamma}{N} \sum_{i\in[d]}	\sum_{n\in A_i} |\nabla f(x)_i| |g_n (x)_i - \nabla f(x)_i |\\
		&\leq \frac{\gamma}{N} \sum_{i\in[d]}	\sum_{n\in A_i} |g_n (x)_i - \nabla f(x)_i |^2 + \frac{\gamma}{N} \sum_{i\in[d]}	\sum_{n\in A_i} |g_n (x)_i | |g_n (x)_i - \nabla f(x)_i |\\
	\left\{\text{Cauchy-Schwarz inequality}\right\}\hspace{10mm}&\leq \frac{\gamma}{N} \sum_{i\in[d]}	\sum_{n\in A_i} |g_n (x)_i - \nabla f(x)_i |^2 + \frac{\gamma}{N} 	\sqrt{\sum_{i\in[d]}\sum_{n\in A_i} |g_n (x)_i |^2} \sqrt{\sum_{i\in[d]}\sum_{n\in A_i}|g_n (x)_i - \nabla f(x)_i |^2}\\
	&\hspace{-50mm}\left\{\sum_{n\in [N]} \left\|g_n(x)\right\|^2 \leq M^2 N,  \sum_{n\in [N]} \left\|g_n(x)_i-\nabla f(x)_i\right\|^2 \leq \sigma_{i}^2 N\right\}\hspace{10mm}\leq \frac{\gamma}{N} \sum_{i\in[d]}	\sigma_{i}^2 N + \frac{\gamma}{N}\sqrt{M^2 N} \sqrt{\sum_{i\in[d]}\sigma_{i}^2 N},
	\end{align*}
	
	Therefore, for $\texttt{err}(GD_2)$ we have  
	\begin{equation}\label{eqn:claim1}
	\texttt{err}(GD_2)\leq \gamma \|\sigma\|^2 + \gamma \|\sigma\|M.
	\end{equation}
	
	We bound $\texttt{err}(signGD_2)$ in a similar vein as in~\citep{signSGD}. Note that
	\begin{align*}
	\texttt{err}(signGD_2) &= \frac{2\gamma c}{N} \sum_{i \in [d]} \sum_{n \in [N]} \nabla f(x)_i \cdot \mathbf{1} \left\{ \texttt{sign}(\nabla f(x)_i) \neq \texttt{sign} (g_n(x)_i) \right\} \\
	&= \frac{2\gamma c}{N} \sum_{i \in [d]} |\nabla f(x)_i|  \sum_{n \in [N]} \mathbf{1} \left\{ |\nabla f(x)_i - g_n(x)_i  | \geq |\nabla f(x)_i |\right\}.
	\end{align*}
	Let $E_i := \{ n \in [N] \colon |\nabla f(x)_i - g_n (x)_i | \geq | \nabla f(x)_i| \}$. Then $\texttt{err}(signGD_2) = \frac{2\gamma c}{N} \sum_{i\in [d]} |\nabla f(x)_i| |E_i|$. Note that
	\[
    \frac{|E_i|}{N} \leq \frac{1}{N}\sum_{n \in [N]} \frac{|\nabla f(x)_i - g_n(x)_i|}{|\nabla f(x)_i|} \leq \frac{1}{|\nabla f(x)_i|}\sqrt{\frac{1}{N}\sum_{n\in[N]} |\nabla f(x)_i - g_n(x)_i|^2} \leq \frac{\sigma_i}{|\nabla f(x)_i|},
	\]
	using the Cauchy-Schwarz inequality. This yields
	\begin{equation}\label{eqn:claim2}
	\texttt{err} (signGD_2) \leq 2\gamma c \|\sigma \|_1.
	\end{equation}

We now want to prove that either $\texttt{err}(GD) \leq (1-\alpha) \gamma \| \nabla f(x)\|^2$ or $\texttt{err}(signGD_2) \leq (1-\alpha) c\gamma \|\nabla f(x) \|_1$. For the sake of contradiction, suppose that 
   \begin{equation}\label{eqnContradiction}
   \texttt{err}(GD) > (1-\alpha) \gamma \| \nabla f(x)\|^2, \hspace{10mm} \texttt{err}(signGD) > (1-\alpha) c\gamma \|\nabla f(x) \|_1.
   \end{equation}
   It is easy to see that these conditions, for $\texttt{err}(signGD)$ imply
   \begin{equation}\label{eqnContradiction1}
    \frac{c\gamma}{N} \sum_{i\in[d]} |\nabla f(x)_i| |B_i| \geq \texttt{err}(signGD) > (1-\alpha) c\gamma \|\nabla f(x) \|_1,
   \end{equation}
   whereas for $\texttt{err}(GD)$ they imply 
   \begin{equation}\label{eqnContradiction2}
\frac{\gamma}{N} \sum_{i\in[d]} |\nabla f(x)_i|^2 |A_i| \geq \texttt{err}(GD) > (1-\alpha) \gamma \|\nabla f(x) \|^2.
\end{equation}  
Let 
\[
A:= \{ i\in [d] \colon |A_i| \geq (1-\alpha) N\}, \hspace{10mm} B:= \{ i \in [d] \colon |B_i| \geq (1-\alpha) N\},
\]
noting that $A\cap B = \emptyset$ and $A\cup B \subseteq [d]$. Moreover, since each $(A_i,B_i)$ is a partition of $[N]$, we have $|B_i|<\alpha N$, for all $i\in A$, and $|A_i| < \alpha N$ for all $i\in B$. Rewriting (\ref{eqnContradiction1}) yields
\begin{align*}
(1-\alpha) c\gamma \sum_{i\in[d]} |\nabla f(x)_i| &< \frac{c\gamma}{N} \sum_{i\in A} |\nabla f(x)_i| |B_i|+\frac{c\gamma}{N} \sum_{i\in B} |\nabla f(x)_i| |B_i|+\frac{c\gamma}{N} \sum_{i\notin A \cup B} |\nabla f(x)_i| |B_i| \\
&< \alpha c\gamma  \sum_{i\in A} |\nabla f(x)_i| +\frac{c\gamma}{N} \sum_{i\in B} |\nabla f(x)_i| |B_i|+(1-\alpha)c\gamma \sum_{i\notin A \cup B} |\nabla f(x)_i| \\
\Longleftrightarrow\hspace{10mm} (1-2\alpha) \sum_{i\in A} |\nabla f(x)_i| &< \sum_{i\in B} |\nabla f(x)_i| \left( \frac{|B_i|}{N} - (1-\alpha)\right) \leq \alpha \sum_{i\in B} |\nabla f(x)_i|.
\end{align*}
It is easy to see that this implies 
\begin{equation}\label{eqnContradiction3}
d\alpha \max_{i\in B} |\nabla f(x)_i| \geq (1-2\alpha)\max_{i\in A}|\nabla f(x)_i|.
\end{equation}
Rewriting (\ref{eqnContradiction2}) in the similar vein yields
\begin{align*}
(1-2\alpha) \sum_{i\in B} |\nabla f(x)_i|^2 &< \alpha \sum_{i\in A} |\nabla f(x)_i|^2,
\end{align*}
which together with \ref{eqnContradiction3} implies
\[
d\alpha \max_{i\in A} |\nabla f(x)_i|^2 > (1-2\alpha)\max_{i\in B}|\nabla f(x)_i|^2 > \frac{(1-2\alpha)^3}{(d\alpha)^2} \max_{i\in A} |\nabla f(x)_i|^2,
\]
which holds only if $\alpha > \frac{1}{d+2}$, contradicting our assumption. Therefore, either $\texttt{err}(GD) \leq (1-\alpha) \gamma \| \nabla f(x)\|^2$ or $\texttt{err}(signGD_2) \leq (1-\alpha) c\gamma \|\nabla f(x) \|_1$, which proves Claim 1.

\vspace{5mm}
Continuing the proof of Theorem~\ref{thmConvergenceAnalysis}, we now bound $\frac{\gamma}{N}\sum_{n\in[N]}I_{n}^{(2)}$ by
\begin{align*}
 \frac{\gamma}{N}\sum_{n\in[N]}I_{n}^{(2)}&=\frac{\gamma}{N} \sum_{n\in [N]} \inn{\nabla f(x_t)}{\texttt{clip}(F_{n}' (x_t),c)-\texttt{clip}(\texttt{top-k}(F_{n}' (x_t)),c)}  \\
 \left\{ \text{Cauchy-Schwarz inequality} \right\}\hspace{10mm}&\leq\gamma \|\nabla f(x_t)\| \frac{1}{N}\sum_{n\in[N]}\|\texttt{clip}(F_n'(x_t),c) - \texttt{clip}(\texttt{top-k}(F_n'(x_t)),c)\|.
 \end{align*}

We will now bound the RHS in two different ways. First, by expanding different cases per coordinate, it is easy to see that
\begin{align*}
 \|\texttt{clip}(F_n'(x_t),c) - \texttt{clip}(\texttt{top-k}(F_n'(x_t)),c)\| \leq \|F_n'(x_t) -\texttt{top-k}(F_n'(x_t))\| \leq \tau_k \|F_n' (x_t)\|,
\end{align*}
by assumption. Therefore,
\begin{align*}
 \frac{\gamma}{N}\sum_{n\in[N]}I_{n}^{(2)}&\leq \gamma \|\nabla f(x_t)\| \tau_k\frac{1}{N} \sum_{n\in[N]}\|F_n'(x_t)\| \leq \gamma \tau_k M^2,
\end{align*}
by the Cauchy-Schwarz inequality and the assumption $\frac{1}{N}\sum_{n\in[N]}\|F_n'(x_t)\|^2 \leq M^2$.

On the other hand, since $\|\texttt{clip}(F_n'(x_t),c) - \texttt{clip}(\texttt{top-k}(F_n'(x_t)),c)\|\leq c(d-k)$, we easily get
\begin{align*}
 \frac{\gamma}{N}\sum_{n\in[N]}I_{n}^{(2)}&\leq \gamma M \min \{\tau_k M, c(d-k)\}.
\end{align*}

Combining the bounds with respect to $I_{n}^{(1)}$ and $I_{n}^{(2)}$ and adding easy analysis of different cases for scaling by $c$ yields
\begin{equation}\label{eqn:boundI}
-\frac{\gamma}{N} \sum_{n\in [N]} I_n (x_t)  \leq -\frac{\gamma\min\{c,1\}}{d+2} \min\{ \| \nabla f(x_t)\|^2 ,  \| \nabla f(x_t) \|_1\} + \gamma \max\{\|\sigma\|^2 + \|\sigma\|M,2 \|\sigma\|_1\}+ \gamma  \min \{\tau_k M^2, c(d-k)M\}.
\end{equation}
}

For $J_n(x_t)$ note that 
	\begin{align*}
	\frac{L\gamma^2}{N} \sum_{n\in [N]} J_n(x_t) &= \frac{L\gamma^2}{N}\sum_{n\in [N]}\E_{\xi_t}\left\| Q({\kk\texttt{clip}}(\texttt{top-k}(F_n'(x_t)),c),\xi_t) \right\|^2\\
	\left\{ \|a+b\|^2 \leq 2\|a\|^2 + 2\|b\|^2\right\} \hspace{10mm} &\leq \frac{2L\gamma^2}{N}\sum_{n\in [N]}\E_{\xi_t}\left\| Q({\kk\texttt{clip}}(\texttt{top-k}(F_n'(x_t)),c),\xi_t) - {\kk\texttt{clip}}(\texttt{top-k}(F_n'(x_t)),c) \right\|^2 \\
	&\hspace{20mm}+\frac{2L\gamma^2}{N} \sum_{n\in[N]} \|{\kk\texttt{clip}}(\texttt{top-k}(F_n'(x_t)),c)\|^2 \\
	\left\{ \E_\xi \left[ \| Q(x,\xi) - x\|^2 \right] \leq \tilde{\sigma}^2\right\} \hspace{10mm}&\leq \frac{2L\gamma^2}{N} \sum_{n\in[N]}\left(\tilde \sigma^2 + \|{\kk\texttt{clip}}(\texttt{top-k}(F_n'(x_t)),c)\|^2 \right)\\
	\left\{ \frac{1}{N} \sum_{n\in [N]} \left\|F_n'(x)\right\|^2 \leq M^2\right\} \hspace{10mm}&\leq 2L\gamma^2 (\tilde \sigma^2 + {\kk\min\{c^2,M^2\}}).
	\end{align*}

Combining bounds on $I_n(x_t)$ and $J_n(x_t)$ yields
\begin{align*}
&{\kk\frac{\gamma\min\{c,1\}}{d+2} \min\{ \| \nabla f(x_t)\|^2, \| \nabla f(x_t) \|_1\}} \leq f(x_t) - \E_\cN \E_{\xi_t} f(x_{t+1}) \\
&\hspace{10mm}+ {\kk\gamma \max\{\|\sigma\|^2 + \|\sigma\|M,2 \|\sigma\|_1\}}+ \gamma  \min \{\tau_k M^2, c(d-k)M\} + 2L\gamma^2 (\tilde \sigma^2 + \min \{c^2,M^2\}) + L\gamma^2 Ak.
\end{align*}

Summing over all $t\in[T]$ yields
\begin{align*}
&{\kk\frac{\gamma\min\{c,1\}}{d+2}\sum_{t\in [T]} \min\{ \E \| \nabla f(x_t)\|^2, \E \| \nabla f(x_t) \|_1\}} \leq f(x_0) - f(x^*) \\
&\hspace{10mm}+ T\gamma \big(   \min \{\tau_k M^2, c(d-k)M\} +  L\gamma Ak + {\kk\max\{\|\sigma\|^2 + \|\sigma\|M,2 \|\sigma\|_1\}}  + 2L\gamma (\tilde \sigma^2 + \min \{c^2,M^2\}) \big),
\end{align*}
which after dividing with $T\gamma $ finishes the proof of the first part.

For the moreover part, in which no quantization is performed, \ie, $Q(x,\xi)=x$, for all $x$ (point-wise, not just on average), the calculation for $J_n(x_t)$ becomes
\begin{align*}
\frac{L\gamma^2}{N} \sum_{n\in [N]} J_n(x_t) &= \frac{L\gamma^2}{N}\sum_{n\in [N]}\left\| {\kk\texttt{clip}}(\texttt{top-k}(F_n'(x_t)),c) \right\|^2 \leq L\gamma^2 {\kk\min\{c^2,M^2\}}.
\end{align*}
Continuing the proof as above (summing over all $t \in [T]$ and dividing by $T\gamma$) finishes the proof of the second part.

\end{document}

%% file: macros.tex
\usepackage{amsthm}
\usepackage{algorithm}
\usepackage{algorithmicx}
\usepackage{algpseudocode}

\theoremstyle{definition}
\newtheorem{definition}{Definition}
\theoremstyle{plain}
\newtheorem{theorem}{Theorem}
\theoremstyle{plain}
\newtheorem{lemma}{Lemma}

\newcommand\Range[1]{\mathrm{Range}(#1)}

\usepackage{graphicx}
\usepackage{caption}
\usepackage{subcaption}

\usepackage{color}

\usepackage{bm}

\newcommand{\m}[1]{{\textcolor{black}{#1}}}
\newcommand{\mo}[1]{{\textcolor{black}{#1}}}
\newcommand{\kk}[0]{\color{black}}

\usepackage{float}
\usepackage{etoolbox,siunitx}
\robustify\bfseries
\usepackage{booktabs}
\usepackage{collcell}
\usepackage{makecell}
\usepackage{supertabular}

\usepackage{array}
\newcommand{\PreserveBackslash}[1]{\let\temp=\\#1\let\\=\temp}
\newcolumntype{C}[1]{>{\PreserveBackslash\centering}p{#1}}
\newcolumntype{R}[1]{>{\PreserveBackslash\raggedleft}p{#1}}
\newcolumntype{L}[1]{>{\PreserveBackslash\raggedright}p{#1}}

\newcommand{\sysname}{\textsc{DataLens}\xspace}
\newcommand{\name}{DataLens\xspace}
\newcommand{\dtwop}{D$^2$P-\textsc{Fed}\xspace}
\newcommand{\fetch}{FetchSGD\xspace}

\newcommand{\ie}[0]{\textit{i.e.}}
\newcommand{\eg}[0]{\textit{e.g.}}
\newcommand{\eps}[0]{\varepsilon}
\renewcommand{\top}[0]{\textsc{TopAgg}\xspace}
\newcommand{\gmdp}[0]{\textsc{GM-DP}\xspace}

\renewcommand{\bf}[1]{\mathbf{#1}}
\newcommand{\cal}[1]{\mathcal{#1}}
\newcommand{\norm}[1]{\lVert{#1}\rVert}
\renewcommand{\rm}[1]{\mathrm{#1}}

\newcommand{\RR}{\mathbb{R}}

\newcommand{\E}{\mathbb{E}}

\newcommand{\inn}[2]{\langle #1, #2 \rangle}

\def \cN {\mathcal N}

%% file: abstract.tex
Recent success of deep neural networks (DNNs) 
hinges on the availability
of large-scale dataset; however, training on such dataset
 often poses privacy 
risks for sensitive training information.
In this paper, we aim to explore the power of generative models and gradient sparsity, and propose a scalable privacy-preserving generative model \sysname, which is able to generate synthetic data in a differentially private (DP) way given sensitive input data. Thus, it is possible to train models for different down-stream tasks with the generated data while protecting the private information. In particular, we leverage the generative adversarial networks (GAN) and PATE framework to train multiple discriminators as ``teacher" models, allowing them to vote with their gradient vectors to guarantee privacy.

Comparing with the standard PATE privacy preserving framework which allows teachers to vote on \textit{one-dimensional} predictions, voting on the \textit{high dimensional gradient vectors} is challenging in terms of privacy preservation. 
As dimension reduction
techniques are required, we need to 
navigate a delicate
tradeoff space between
(1) the improvement 
of privacy preservation
and (2) the slowdown of 
SGD convergence.
To tackle this, we 
propose a novel dimension compression and aggregation approach \top, which combines
 top-$k$ dimension compression with
a corresponding noise injection mechanism.
We theoretically prove that the \sysname framework guarantees differential privacy for its generated data,
and  provide a novel analysis on its convergence 
\m{
to illustrate such a tradeoff on privacy and convergence rate, which requires non-trivial analysis as it requires a joint analysis on gradient compression, coordinate-wise gradient clipping, and DP mechanism.}
To demonstrate the practical usage of \sysname, we conduct extensive experiments on diverse datasets including MNIST, Fashion-MNIST, and high dimensional CelebA and Place365 datasets. We show that \sysname significantly outperforms other baseline differentially private data generative models.
\mo{Our code is publicly available at \url{https://github.com/AI-secure/DataLens}.}


%% file: introduction.tex
Advanced machine learning methods, especially deep neural networks (DNNs), have achieved great success in a wide array of applications~\citep{Jia2019EfficientTD, Jia2019TowardsED, Ye2020ReinforcementLearningBP}, mainly due to the fast development of hardware, their expressive representation power, and the availability of large-scale training datasets.
However, one major concern that has risen in machine learning is that the training data usually contain a large amount of privacy sensitive information (e.g., human faces and medical records), which could be leaked via the trained machine learning models~\citep{membershipAttack, zhang2020secret}.
\textit{How to protect such private information while allowing high learning utility for the dataset} has attracted a lot of attention. Differentially private (DP) deep learning~\citep{Abadi2016DeepLW} proposes adding Gaussian noise to the clipped gradient during training, thus ensuring that the learned results are differentially private regarding the training data. However, its learning utility largely decreases with strong privacy requirements.
A semi-supervised learning framework PATE~\citep{papernot2016semi, papernot2018scalable} is later proposed to improve the learning effectiveness at the presence of privacy noise, by leveraging the aggregation of noisy teacher models trained on private datasets. It is shown that the PATE framework is able to improve the learning utility significantly while protecting data privacy.
\m{However, applying such privacy preserving training framework from the discriminative model to the generative model to guarantee that the generated data is differentially private is non-trivial given the potential high-dimensional gradient aggregation.}

To further improve the flexibility of differentially private machine learning process, in this paper we aim to design a \textit{privacy-preserving data generative model} which ensures that the data generator and the generated data, instead of only the predictions, are differentially private. This way, the generated data can then be used to train arbitrary models for different down-stream tasks with high flexibility. Having in mind that the generative adversarial networks (GAN) \citep{goodfellow2014generative} achieved great success in terms of generating high quality data, it is natural to ask: \textit{Is it possible to leverage the power of GAN in a way to generate data in a differentially private manner?}
Some recent works have shown promising results on differentially private data generative models~\citep{long2019scalable, yoon2018pategan}.
However, most of them can only generate low dimensional data such as tabular data with weak privacy guarantees (i.e., $(\epsilon, \delta) - DP$ with small $\epsilon$).
The problem of \textit{generating differentially private high dimensional data} (e.g., image) with strong privacy guarantees is still open, due to the fact that, in order to achieve strong privacy guarantees, the limited privacy budget is not enough to train a generative model to approximate any high dimensional perturbation.

In the meantime, an independent line of research
concerning gradient compression in distributed 
training for communication efficiency~\citep{10.5555/3326943.3327063,NIPS2018_7837,DBS-062,pmlr-v80-bernstein18a} shows that some
noisy
compression schemes such as only keeping the top-$K$ elements of the gradient would achieve 
statistically similar
convergence rate with vanilla training. This observation could potentially 
be a remedy for the above problem of high dimensionality  --- Intuitively, the noises introduced by 
these noisy compression schemes could also help 
protect privacy and combining them with 
traditional DP noise mechanism may allow us to add
fewer amount of noise to achieve the same level of DP protection.
This intuition inspired our work, which, to our best knowledge,
is the \textit{first} to marry these two lines of research on privacy and communication-efficient distributed learning
to achieve both differential privacy guarantees and 
high model utility on high-dimensional data. As we will see, though intuitively feasible, taking advantage of this intuition is far from trivial.

Specifically, we propose a differentially private data generative model \sysname based on the PATE framework, which trains multiple discriminators as different \textit{teacher} models to provide the back-propagation information in a differentially private way to the \textit{student} generator. In addition, to tackle the high-dimensional data problem we mentioned above, we propose an effective noisy gradient compression and aggregation strategy \top to allow each discriminator to vote for the top several dimensions in their gradients and then aggregate their noisy gradient sign to perform back-propagation. We prove the differential privacy guarantees for both the data generator and generated data for \sysname. 
Furthermore, to ensure the performance of the trained DP generative model,
we provide a theoretical \textit{convergence} analysis for the proposed gradient compression and aggregation strategy. 
\m{In particular, to our best knowledge, this is the  first convergence analysis considering the \textit{coordinate-wise} gradient clipping together with gradient compression and DP noise mechanism.}

Finally, we conduct extensive empirical evaluation on the utility of the generated based on \sysname comparing with several other baselines on image datasets such as MNIST, Fashion-MNIST,  CelebA, and Place365, which is of much higher dimension than the tabular data used by existing DP generative models. We show that the generated data of \sysname can achieve the state-of-the-art utility on all datasets compared with baseline approaches. \m{We also conduct a series ablation studies to analyze the visualization quality of the generated data, the data-dependent and data-independent privacy bounds, the impact of different components and hyper-parameters in \sysname, as well as different gradient compression methods. }

In addition, to further evaluate the proposed compression and aggregation strategy \top, which is the key building block in  \sysname, we also discuss and evaluate \top for the standard DP SGD training. We show that on both MNIST and CIFAR-10 datasets, \top can achieve similar or even better model utility than the state of the art baseline approaches, which leads to an interesting future direction.

{\bfseries \underline{Technical Contributions.}}
\m{In this paper, we propose a general and effective differentially private data generative model for high-dimensional image data.
We make contributions on both theoretical
and empirical front.}

\begin{itemize}[leftmargin=*]
\item We propose an effective differentially private data generative model \sysname, which can be applied for generating high-dimensional image data with limited privacy budgets.
\item We prove the privacy guarantees for \sysname, and conduct thorough theoretical analysis for the convergence of \sysname. We show that \sysname is able to make a good tradeoff between the privacy protection by adding DP noise and the slowdown of SGD convergence due to the added DP noise.
\item We propose a novel noisy gradient compression and aggregation algorithm \top by combining the top-$k$ dimension compression and a specific DP noise injection mechanism.
We also discuss the potential of adapting  \top to standard DP SGD training with evaluations.
\item \m{To  illustrate tradeoff between
differential privacy and convergence given gradient compression, we provide a novel theoretical analysis
jointly considering gradient compression, coordinate-wise gradient clipping, and DP mechanism.}
\item We conduct extensive empirical evaluation on \sysname with four image datasets, including MNIST, Fasion-MNIST,  CelebA, and Place365 datasets.
We show that in term of the utility of generated data, \sysname significantly outperforms the state-of-the-art DP generative models. 
\end{itemize}

%% file: preliminaries.tex
\vspace{-2mm}
\subsection{Differential Privacy}



$(\varepsilon,\delta)$-differential privacy ($(\varepsilon,\delta)$-DP) is currently an industry standard of privacy notion proposed by \citet{dwork2008differential} 
It bounds the change in output distribution caused by a small input difference for a randomized algorithm. The following definition formally describes this privacy guarantee.
\begin{definition}[$(\varepsilon, \delta)$-Differential Privacy~\citep{dwork2008differential}]
\label{def:dp}
A randomized algorithm $\mathcal{M}$ with domain $\mathbb{N}^{|\mathcal{X}|}$ is $(\varepsilon,\delta)$-differentially private if for all $\mathcal{S} \subseteq \Range{\mathcal{M}}$ 
and for any neighboring datasets $D$ and $D'$:
\begin{equation*}
    \Pr[\mathcal{M}(D) \in \mathcal{S}] \leq \exp(\varepsilon)\Pr[\mathcal{M}(D')\in \mathcal{S}] + \delta.
\end{equation*}
\end{definition}

Differential privacy is immune to post-processing. Formally, the composition of a data-independent mapping $g$ with an $(\varepsilon, \delta)$-DP mechanism $\mathcal{M}$ is also $(\varepsilon, \delta)$-DP~\citep{dwork2014algorithmic}.

\paragraph{PATE Framework} Private Aggregation of Teacher Ensembles (PATE) is one of the DP mechanisms~\citep{papernot2016semi, papernot2018scalable} 
that provide the differential privacy guarantees for trained machine learning models. The PATE framework achieves DP by aggregating the prediction votes from several \textit{teacher models}, which are trained on private data, as the input with DP noise for a \textit{student model}, which serves as the final released prediction model with privacy protection.
The privacy analysis \citep{papernot2016semi}  of PATE is derived using Laplacian mechanism and moments accountant technique based on \citet{Abadi2016DeepLW}, which yields a tight privacy bound when the outputs of teacher models have high consensus over the topmost votes. 


\vspace{-2mm}
\subsection{Data Generative Models}
\label{sec:related-gen}
Data generative models aim to approximate the distribution of large datasets and thus generate diverse datasets following the similar data distribution., which can be used for data augmentation and further analysis. Recently, Generative Adversarial Network (GAN) \citep{goodfellow2014generative} has been proposed as a deep learning architecture for training generative models.
In particular, GAN consists of a generator ${\Psi}$ that learns to generate synthetic records, and a discriminator $\Gamma$ that is trained to tell real records apart from the fake ones. Given an input datase $x$ and a sampled noise $z$, we train the discriminator $\Gamma$  to minimize the likelihood of classifying the synthetic example  from ${\Psi}$ as drawing from the real distribution with the loss function $\mathcal{L}_\Gamma$ defined as:
$
    \mathcal{L}_\Gamma = - \log {\Gamma}(x) - \log (1- {\Gamma}({\Psi}(z)).
    $
The generator $\Psi$ seeks to maximize the probability of the generated data being predicted as real ones by the discriminator $\Gamma$ with the loss function $\mathcal{L}_{\Psi}$ defined as:
$
    \mathcal{L}_\Psi = - \log {\Gamma}(\Psi(z)).
    $
Though GANs are able to generate high-quality data records given large training datasets, such generative models are prone to leak the information of training data \citep{ChenYZF20}. This presents us the challenge on \textit{how to prevent the training information leakage for generated data}, especially when the training data contains a large amount of privacy-sensitive information. In this paper, we aim to train differentially private generative models so that we can enjoy the benefits of generative models to generate unlimited amount of high-utility data for arbitrary downstream tasks, while protecting sensitive training  information.


\vspace{-4mm}
\m{\subsection{Gradient Compression}}
\label{sec:related-compression}

\m{
Gradient compression
techniques, such as quantization, low-rank approximation, and sparsification, have been 
studied in the last 
decade~\citep{Koloskova2019-kl, Tang2018-ec, Chen2020-mm, Bernstein2018-mv, Lim2019-hk, Vogels2020-wd}. One surprising result 
is that stochastic gradient descent 
are often robust to these 
operations --- one can often 
compress the data by orders
of magnitude without 
significantly slow down
the convergence. Most of existing 
efforts focus on saving 
the communication overheads 
in distributed training. 
}

\m{
This paper is inspired by these 
previous research, however, 
focuses on a different problem --- \textit{can the saving of communication overheads
provide benefits to 
differential privacy?} As we will see,
by compressing the gradient in certain way, we are able to decrease the 
dimension of the gradient without significantly slow down the convergence. This can translate into fewer amount of noises that one needs to add to ensure DP. This intuition, however,
requires careful design of the underlying algorithm and imposes novel 
challenges in theoretical understandings, which is the focus of this work.
}


%% file: algorithm.tex
We now present the main algorithms used in \sysname. It consists of three parts: an ensemble of teacher discriminators, a student generator, and a DP gradient aggregator. First, we introduce the algorithm for training the student generator and teacher discriminators. Then, we introduce the novel high-dimensional DP gradient compression and aggregation algorithm \top (\Cref{algo:gradAgg}). This algorithm consists of two parts: a top-$k$ gradient compression algorithm (\texttt{TopkStoSignGrad},~\Cref{algo:topk}) that compresses the gradient vectors while preserving the important gradient directions; and a DP gradient aggregation algorithm that aggregates teacher gradient vectors with differential privacy guarantees. 

\textbf{Training DP Generator via Teacher Discriminator Aggregation.}
\m{On the high level, as shown in Figure~\ref{fig:overview} the teacher discriminators are trained on non-overlapping sensitive data partitions to distinguish between real and synthetic data. The student generator produces synthetic records, sends them to the teachers for label querying, and uses the aggregated gradient from the teacher discriminators to improve its generated synthetic records.} The DP gradient aggregator ensembles the teachers' gradient vectors and adds DP noise for  privacy guarantees. The detailed algorithm for this process is included in the~\Cref{algo:trainGen}. 

To begin with, we randomly partition the sensitive training dataset into non-overlapping subsets of the same size. Each partition is associated with one teacher discriminator. Then, we iteratively update the student generator and the teacher discriminators. Each iteration consists of the following four steps:

\label{sec:algo}
\begin{small}
\begin{algorithm}[t]
\caption{\textbf{- Training the Student Generator.}} 
\label{algo:trainGen}
\begin{algorithmic}[1]
    \State {\bfseries Input:} batch size $m$, number of teacher models $N$, number of training iterations $T$, gradient clipping constant $c$, top-$k$, noise parameters $\sigma$, voting threshold $\beta$, disjoint subsets of private sensitive data $d_1, d_2, \dots, d_N$, learning rate $\gamma$
    \For{number of training iterations $\in [T]$}
        \State \Comment{Phase I: Pre-Processing}
        \State Sample $m$ noise samples  $\mathbf{z} =\left(\mathbf{z}_1, \mathbf{z}_2, \dots, \mathbf{z}_m\right)$
        \State Generate fake samples $\Psi(\mathbf{z}_1), \Psi(\mathbf{z}_2), \dots, \Psi(\mathbf{z}_m)$ 
        \For{each synthetic image $\Psi(\mathbf{z}_j)$}
            \State \Comment{Phase II: Private Computation and Aggregation}
            \For{each teacher model ${\Gamma}_i$}
                \State Sample $m$ data samples { $\left(\bf x_1, \bf x_2, \dots, \bf x_m\right)$} from $d_i$
                \State Update the teacher discriminator $\Gamma_i$ by descending its stochastic gradient on $\mathcal{L}_{\Gamma_i}$ on both fake samples and real samples
                \State \m{Calculate the  gradient $\mathbf{g}_j^{(i)}= \left. -\frac{\partial \log {\Gamma_i}(a)}{\partial a} \right |_{a=\Psi(\mathbf{z}_j)}$  of the teacher discriminator loss $\mathcal{L}_{\Gamma_i}$  w.r.t. the sample $\Psi(\mathbf{z}_j)$}.
            \EndFor
            \State $\mathbf{g}_j \leftarrow (\mathbf{g}_j^{(1)}, \mathbf{g}_j^{(2)}, \dots, \mathbf{g}_j^{(N)})$
            \State $\mathbf{\bar{g}}_j \leftarrow \texttt{DPTopkAgg}\left(T, \mathbf{g}_j, c, k, \sigma, \beta \right)$ 
            \State \Comment{Phase III: Post-Processing}
            \State $\mathbf{\hat{x}}_j \leftarrow \Psi(\mathbf{z}_j) + \gamma \m{\mathbf{\bar{g}}}_j $ 
        \EndFor
        \State Update the student generator $\Psi$ by descending its stochastic gradient on \m{ $\hat{\mathcal{L}_{\Psi}}(\mathbf{z},{\mathbf{\hat{x}}}) = \frac{1}{m} \sum_{j=1}^{m} (\Psi(\mathbf{z}_j) - \hat{\mathbf{x}}_j)^2$} on $\mathbf{\hat{x}} =\left(\mathbf{\hat{x}}_1, \mathbf{\hat{x}}_2, \dots, \mathbf{\hat{x}}_m\right)$
    \EndFor
\end{algorithmic}
\end{algorithm}
\end{small}
\textbf{Step 1: Training teacher discriminators.} The student generator $\Psi$ produces a batch of synthetic records. Each teacher discriminator $\Gamma_i$ updates the weights based on standard discriminator loss $\mathcal{L}_{\Gamma_i}$ to reduce its loss on distinguishing the synthetic records from real records in its training data partition.

\textbf{Step 2: Generating and compressing teacher gradient vectors.} 
\m{Each teacher discriminator $\Gamma_i$ computes a gradient vector $g^{(i)}$ of the discriminator loss $\mathcal{L}_{\Gamma_i}$ with regard to the synthetic records.
Such gradient vector contains the information that could guide the student generator to improve its synthetic records aiming to increase the generated data utility (\ie, classification accuracy of trained models). }

\textbf{Step 3: DP gradient compression and aggregation.} 
\m{In order to perform efficient DP mechanism for the teacher gradient vectors, we propose \top to compress  the teacher gradient vectors first and then aggregate them.}
We perform gradient aggregation over the teachers' gradient vectors with a corresponding noise injection algorithm that guarantees differential privacy. The final aggregated noisy gradient vector is then passed to the student generator. Details will be discussed in the next subsection.

\textbf{Step 4: Training the student generator.} The student generator learns to improve its synthetic records by back-propagating the aggregated DP gradient vectors produced by the teacher ensemble. \m{We define the loss function for the student generator as $\hat{\mathcal{L}}_{\Psi}(\mathbf{z},\hat{\mathbf{x}}) = \frac{1}{m} \sum_{j=1}^{m} (\Psi(\mathbf{z}_j) - \hat{\mathbf{x}}_j)^2$, where $\mathbf{z}_j$ is the noise sample, $\Psi(\mathbf{z}_j)$ is the synthetic data, and $\hat{\mathbf{x}}_j = \Psi(\mathbf{z}_j) + \gamma \mathbf{\bar{g}}_j$ is the synthetic data plus the aggregated DP gradient vectors from the teacher discriminators. Since $-\frac{\partial \hat{\mathcal{L}}_{\Psi}(\mathbf{z},\hat{\mathbf{x}})}{\partial \hat{\mathbf{x}}} = \frac{2\gamma}{m} \sum_{j=1}^{m} \mathbf{\bar{g}}_j$, descending the stochastic gradient on $\hat{\mathcal{L}}_{\Psi}(\mathbf{z},\hat{\mathbf{x}})$ would propagate the aggregated DP gradient vectors from the teacher discriminators to the student generator. }


\textbf{Top-$k$ Gradient Compression via Stochastic Sign Gradient.}
\m{In the Step 3. \textit{gradient compression and aggregation}, each teacher model compresses its dense, real-valued gradient vector into a sparse sign vector with $k$ nonzero entries.
We first present and discuss the 
gradient compression 
function:}
$\texttt{TopkStoSignGrad}(\mathbf{g}, c, k) $ (\Cref{algo:topk}).

\m{Inspired by the 
recent results on signSGD~\citep{pmlr-v80-bernstein18a}
and gradient
compression in communication efficient distributed learning~\citep{NIPS2018_7837,10.5555/3326943.3327063}, we design a gradient compression algorithm that reduces a gradient vector in two steps.}
First, we select the top-$k$ dimensions
in each teacher gradient $\mathbf{g}$ and set the remaining dimensions to zero.  This step reduces the dimensionality of the gradient vector and allows us to achieve a tighter privacy bound during DP gradient aggregation. Then, we clip the gradient at each dimension with threshold $c$, normalize the top-$k$ gradient vector, and perform stochastic gradient sign quantization. \m{Specifically, we first select the top-$k$ dimensions of the gradient. Let $\hat{{g}}_j$ be the j-\textit{th} dimension selected from the  gradient vector g,  we then clip each selected dimension as $\hat{g}_j = \min(\max(\hat{g}_j, -c), c)$. After normalization, we 
assign the stochastic gradient sign ${\tilde{g}}_j$ based on the following rule:}
\begin{small}
\begin{equation}
    {\tilde{g}}_j =\left\{
                \begin{array}{ll}
                  1, & \textit{with probability } \frac{1 + \hat{{g}}_j}{2} ;\\
                  -1, & \textit{with probability } \frac{1 - \hat{{g}}_j}{2} .\\
                \end{array}
              \right.
\end{equation}
\end{small}
We can see that $\tilde{g}_j$
is an unbiased estimator 
of $\hat{{g}}_j$.
As a result, we transform a 
dense, real-valued 
gradient vector into 
a sparsified $\{-1, 0, 1\}$-valued vector, which
allows more effective differentially private 
gradient aggregation.


\begin{small}
\begin{algorithm}[t]
\caption{\textbf{- Gradient Compression on Top-$k$ Dimensions via Stochastic Sign Gradient (\texttt{TopkStoSignGrad}).} This algorithm takes in a gradient vector of a teacher model $\mathbf{g}^{(i)}$ and returns the compressed gradient vector $\mathbf{\tilde{g}}^{(i)}$.}
\label{algo:topk}
\begin{algorithmic}[1]
    \State {\bfseries Input:}  Gradient vector $\mathbf{g}^{(i)}$, gradient clipping constant $c$, top-$k$
    \State $\mathbf{h}^{(i)} \leftarrow $ arg-topk($\lvert {\mathbf{g}}^{(i)}\rvert$, $k$) 
    \Statex \hfill\Comment{the top-$k$ indices of the absolute value of gradient $\hat{\mathbf{g}}^{(i)}$}
    
    \State \m{$\mathbf{g}^{(i)}_j = \min(\max(\mathbf{g}^{(i)}_j, -c), c)$ for each dimension $j$ in ${\mathbf{g}}^{(i)}$
    \Statex \hfill\Comment{Clip each dimension of $\mathbf{g}^{(i)}$ so that $-c \le \mathbf{g}_j^{(i)} \le c.$} }
    \State $\hat{\mathbf{g}}^{(i)} \leftarrow \mathbf{g}^{(i)} /  \left\|\mathbf{g}^{(i)} \right\|_{\infty}$ \quad \hfill\Comment{gradient normalization to (-1, 1)} 
    \State $\mathbf{\tilde{g}}^{(i)} \leftarrow \mathbf{0}$
    \hfill\Comment{initialization of the compressed sparse gradient vector}
    \For{each top-$k$ index $j$ in $\mathbf{h}^{(i)}$} 
        \State ${\tilde{g}}_j^{(i)} =\left\{
                \begin{array}{ll}
                  1, & \textit{with probability } \frac{1 + \hat{{g}}_j^{(i)}}{2}\\
                  -1, & \textit{with probability } \frac{1 - \hat{{g}}_j^{(i)}}{2}\\
                \end{array}
              \right.$
    \EndFor
    \State {\bfseries Return:} $\mathbf{\tilde{g}}^{(i)}$
\end{algorithmic}
\end{algorithm}
\end{small}

{\textbf{High Dimensional DP Gradient Aggregation.}} 
In the gradient aggregation step, we perform differentially private aggregation on the compressed teachers' gradient vectors. Specifically, we want to guarantee that the change of any teacher gradient vector will not considerably shift the output distribution of the aggregation. \Cref{algo:gradAgg} presents the aggregation algorithm. 

After compression, each gradient vector is a sparse sign vector with $k$ nonzero entries. Therefore, we propose a novel algorithm that converts gradient aggregation into a voting problem. Specifically, the gradient signs can be viewed as votes for the gradient directions. Each teacher can vote for $k$ gradient dimensions. For each dimension in the top-$k$ selection, they vote either the positive direction (\ie, $\tilde{g}_j=1$) or the negative direction (\ie, $\tilde{g}_j=-1$). 

We apply Gaussian mechanism~\citep{mironov2017renyi} with post-processing thresholding to aggregate the gradient votes. First, we take the sum of the gradient vectors and inject Gaussian noise following distribution $\mathcal{N}(0, \sigma^2)$. Then, we check whether the noisy vote for each gradient direction is greater than a threshold. This thresholding step guarantees that we only select the gradient directions with high agreement rate among the teacher models. 
To reach an agreement, the following two conditions need to be satisfied.
First, the gradient dimension is ranked as top-$k$ for the majority of the teachers. Second, these teachers also agree on the sign of the gradients along these dimensions. With thresholding, we remove the influence of outliers among the teachers. Intuitively, since the selected directions have higher votes, they are unlikely to be changed by the DP noise injection mechanism to preserve utility. 

\m{In particular, the Top-$k$ stochastic sign gradient quantization and DP gradient aggregation approaches together form a novel DP gradient compression and aggregation algorithm \top (\Cref{algo:gradAgg}), which serves as a key building block in \sysname. These joint operators are the first time to be adopted in a data generated model, and we will  provide the convergence analysis for these joint operators in Section~\ref{sec:convergence}.}

\begin{small}
\begin{algorithm}[t]
\caption{\textbf{- Differentially Private Gradient Compression and Aggregation (\top).} This algorithm takes gradients of teacher models and returns the compressed and aggregated differentially private gradient vector.}
\label{algo:gradAgg}
\begin{algorithmic}[1]
    \State {\bfseries Input:} Teacher number $N$, gradient vectors of teacher models $\mathcal{G}=\{\mathbf{g}^{(1)}, \dots, \mathbf{g}^{(N)}\}$, gradient clipping constant $c$, top-$k$, noise parameters $\sigma$, voting threshold $\beta$
    \State\Comment{Phase I: Gradient Compression}
    \For{each teacher's gradient $\mathbf{g}^{(i)}$}
        \State $\mathbf{\tilde{g}}^{(i)} \leftarrow \texttt{TopkStoSignGrad}(\mathbf{g}^{(i)}, c, k) $
    \EndFor
    \State\Comment{Phase II: Differential Private Gradient Aggregation}
    \State $\mathbf{\tilde{g}}^* \leftarrow \sum_{i=1}^N {\mathbf{\tilde{g}}^{(i)}}  + \mathcal{N}(0, \sigma^2)$
    \State\Comment{Phase III: Gradient Thresholding (Post-Processing)}
    \For{each dimension ${\tilde{g}}_j^*$ of $\mathbf{\tilde{g}}^*$}
        \State ${{\bar{g}}}_j =\left\{
                \begin{array}{ll}
                  1, &  \text{if} \quad  {\tilde{g}}_j^* \ge \beta N ;\\
                  -1, & \text{if} \quad  {\tilde{g}}_j^* \le -\beta N ;\\
                  0, & \text{otherwise.}
                \end{array}
              \right.$
    \EndFor
    \State {\bfseries Return:} $\mathbf{\bar{g}}$
\end{algorithmic}
\end{algorithm}
\end{small}

%% file: data_independent_privacy_bound.tex
In our PATE based data generative framework, the teacher discriminators have access to the sensitive training data and the student generator learns about the sensitive data from the teachers through the gradient aggregation algorithm. Therefore, if the gradient aggregation algorithm preserves DP or RDP, the same privacy guarantee applies to the student generator based on the post-processing theorems. Hence, we focus on deriving the privacy bound for the gradient aggregation algorithm (\texttt{\top}). 


Let $\mathcal{\tilde{G}}=(\mathbf{\tilde{g}}^{(1)}, \dots, \mathbf{\tilde{g}}^{(N)})$ be the set of compressed teacher gradient vectors, where $\mathbf{\tilde{g}}^{(i)}$ is the compressed gradient of the $i$-th teacher. We define sum aggregation function 
\begin{small}
\begin{equation*}
f_{\rm sum}(\mathcal{\tilde{G}}) = \sum_{i=1}^{N} {\mathbf{\tilde{g}}^{(i)}},
\end{equation*}
\end{small}
and, by applying Gaussian mechanism, we have
\begin{small}
\begin{align*}
    \mathbf{\tilde{G}}_{\sigma} f_{\textrm sum}(\mathcal{\tilde{G}}) 
    = f_{\textrm sum}(\mathcal{\tilde{G}}) + \mathcal{N}\left(0, \sigma^{2}\right)
    = \sum_{\mathbf{\tilde{g}} \in \mathcal{\tilde{G}}} {\mathbf{\tilde{g}}} + \mathcal{N}\left(0, \sigma^{2}\right).
\end{align*}
\end{small}
For any real-valued function $f$, the Gaussian mechanism provides the following RDP guarantee: 
\begin{theorem}[RDP Guarantee for Gaussian Mechanism~\citep{mironov2017renyi}]
\label{theorem:rdp-gaussian}
If $f$ has $\ell_2$-sensitivity $s$, then the Gaussian mechanism $\mathbf{G}_{\sigma}f$ satisfies $\left(\lambda, s^2 \lambda /\left(2 \sigma^{2}\right)\right)$-RDP.
\end{theorem}

Thus, to calculate the RDP guarantee for $\mathbf{\tilde{G}}_{\sigma} f_{\textrm sum}(\mathcal{\tilde{G}})$, we first need to calculate the $\ell_2$ sensitivity~\citep{dwork2008differential} of the aggregation algorithm.

\begin{lemma}
\label{lemma:l2}
For any neighboring top-$k$ gradient vector sets $\mathcal{\tilde{G}}$, $\mathcal{\tilde{G'}}$ differing by the gradient vector of one teacher, the $\ell_2$ sensitivity for $f_{\rm sum}$ is $2\sqrt{k}$.
\end{lemma}

\vspace{-0.5em}
\begin{proof}
The $\ell_2$ sensitivity is the maximum change in $\ell_2$ norm caused by the input change. 
For each of the top-$k$ dimension, a teacher could take one of the following two changes: (1) vote for the opposite direction, which flips the gradient sign of one entry; (2) vote for a different dimension, which reduces the vote of one entry and increases the vote on another. The former changes $\ell_2$ norm by $2$, and the latter by $\sqrt{2}$.  In the worst case, the teacher flips all the top-$k$ gradient signs, the change in $\ell_2$ norm equals $\sqrt{2^2k} = 2\sqrt{k}$.
\end{proof}


\begin{theorem}
\label{theorem:rdp_guarantee}
\m{The \top algorithm (Algorithm~\ref{algo:gradAgg}) guarantees $(\frac{ 2k\lambda}{\sigma^2} +  \frac{\log 1/\delta}{\lambda-1}, \delta)$-differential privacy for all $\lambda \ge 1$ and $\delta \in (0,1)$.}
\end{theorem}

\vspace{-1em}
\begin{proof}
The \texttt{DPTopkAgg} algorithm can be decomposed into applying gradient thresholding on the output of the sum aggregation Gaussian mechanism $\mathbf{G}_{\sigma} f_{\textrm sum}$. 
$\mathbf{G}_{\sigma} f_{\textrm sum}$ guarantees $(\lambda, 2k\lambda/\sigma^2)$-RDP (\Cref{lemma:l2} \& ~\Cref{theorem:rdp-gaussian}), and thus this theorem is the result of applying the post-processing theorem of RDP \m{and~\Cref{theorem:rdp-dp}}. 
\end{proof}

%% file: data_dependent_privacy_bound.tex

The parameters $\varepsilon$ in \Cref{def:dp} and $\alpha$ in \Cref{def:rdp} are called \emph{the privacy budget} of a randomized mechanism. When $\varepsilon$ and $\alpha$ are dependent of the input dataset $D$, the privacy bound is data-dependent. \m{In the following section, we compare the data-independent privacy bound in~\Cref{theorem:rdp_guarantee} with a data-dependent privacy bound proposed by~\citet{papernot2018scalable}. We prove that, when the algorithm has high dimensional outputs, the  data-independent privacy bound (\Cref{theorem:rdp_guarantee}) is tighter and achieves better utility.}

\m{First, we revisit the data-dependent RDP bound for randomized algorithms~\citep{papernot2018scalable}:} 

\begin{theorem}[Data-Dependent RDP Bound~\citep{papernot2018scalable}]
\label{theorem:rdp-dependent}
Let $\mathcal{M}$ be a randomized algorithm with $(\mu_1,\alpha_1)-$RDP and $(\mu_2,\alpha_2)-$RDP guarantees and suppose that there exists a likely outcome $\mathbf{\bar{g}}^*$ given
a dataset $D$ and a bound $\Tilde{q}\leq 1$ such that $\tilde{q} \geq \operatorname{Pr}\left[\mathcal{M}(D) \neq \mathbf{\bar{g}}^* \right]$. Additionally, suppose that $\lambda \leq \mu_{1}$ and $\tilde{q} \leq e^{\left(\mu_{2}-1\right) \alpha_{2}} /\left(\frac{\mu_{1}}{\mu_{1}-1} \cdot \frac{\mu_{2}}{\mu_{2}-1}\right)^{\mu_{2}}$. Then, for any neighboring dataset $D'$ of $D$, we have:
\begin{small}
\begin{align*}
D_{\lambda}\left(\mathcal{M}(D) \| \mathcal{M}\left(D^{\prime}\right)\right) \leq \frac{1}{\lambda-1} \log 
&\left((1-\tilde{q}) \cdot \boldsymbol{A}\left(\tilde{q}, \mu_{2}, \alpha_{2}\right)^{\lambda-1} \right. \\
&+ \left. \tilde{q} \cdot \boldsymbol{B}\left(\tilde{q}, \mu_{1}, \alpha_{1}\right)^{\lambda-1} \right),  
\end{align*}
\end{small}where \vspace{-2mm} 
\begin{small}
\begin{equation*}
\boldsymbol{A}\left(\tilde{q}, \mu_{2}, \alpha_{2}\right) \triangleq(1-\tilde{q}) /\left(1-\left(\tilde{q} e^{\alpha_{2}}\right)^{\frac{\mu_{2}-1}{\mu_{2}}}\right),\qquad \boldsymbol{B}\left(\tilde{q}, \mu_{1}, \alpha_{1}\right) \triangleq e^{\alpha_{1}} / \tilde{q}^{\frac{1}{\mu_{1}-1}}.
\end{equation*}
\end{small}
\end{theorem}
The parameters $\mu_1$ and $\mu_2$ are optimized to get a data-dependent RDP guarantee for any order $\lambda$. 

\m{The above data-dependent RDP bound is tighter than the data-independent bound in \Cref{theorem:rdp_guarantee} when $\tilde{q} \ll 1$. Since $\tilde{q} \geq \operatorname{Pr}\left[\mathcal{M}(D)\right.$ $\left.\neq \mathbf{\bar{g}}^* \right]$, the data-dependent bound improves upon the data-independent bound only when the algorithm's output distribution peaks at a likely outcome $\mathbf{\bar{g}}^*$.}
\citet{papernot2018scalable} demonstrated that the  data-dependent privacy bound improves the utility of the PATE framework when teachers vote on one-dimensional predictions. However, we observe that this bound does \emph{not} always guarantee a better utility for algorithms with high dimensional outputs. Specifically, with the increase of the output dimensionality, there is a diminishing benefit from using the data-dependent privacy bound in \Cref{theorem:rdp-dependent}. 

Below, we demonstrate \m{the observation that the data-independent privacy bound can achieve better utility} with the aggregation and thresholding steps in \top. Let $\mathcal{M}(\mathcal{\tilde{G}}, N, \beta)$ represent the composition of these two steps, where $\mathcal{\tilde{G}}$ is the compressed gradient vector set, $N$ is the number of teachers, and $\beta$ is the voting threshold.  


\begin{theorem}
\label{theorem:data_dependent_prob}
For any $\mathbf{\bar{g}}^* \in \{0,1\}^{d}$, we have \vspace{-1mm}
\begin{small}
\begin{align*}
 \Pr[\mathcal{M}(\mathcal{\tilde{G}}, N, \beta) \neq \mathbf{\bar{g}}^*] 
     &= 1 - 
    \prod_{\{j \mid \bar{g}^*_j=1\}}\left(1-\Phi\left(\frac{\beta N-f_j}{\sigma}\right)\right)\\
    &\prod_{\{j \mid \bar{g}^*_j=-1\}}\Phi\left(\frac{\beta N - f_j}{\sigma}\right)
    \prod_{\{j \mid \bar{g}^*_j=0\}}\erf\left(\frac{\beta N - f_j}{\sqrt{2}\sigma}\right)
\end{align*}
\end{small}
where $\Phi$ is the cumulative distribution function of the  normal distribution, $\erf$ is the error function, and $f_j$ is the $j$-th dimension of the gradient vector sum $\sum_{i=1}^N {\mathbf{\tilde{g}}^{(i)}}$ without the noise injection.
\end{theorem}

\Cref{theorem:data_dependent_prob} shows that the bound $\tilde{q} \geq \Pr[\mathcal{M}(\mathcal{\tilde{G}}, N, \beta) \neq \mathbf{\bar{g}}^*]$ increases with the increasing output dimensionality of $\mathcal{M}$. Since the Gaussian mechanism adds independent Gaussian noise along each dimension, this noise flattens out the probability distribution around the likely outcome $\mathbf{\bar{g}}^*$, and consequently reduces the peak probability for $\Pr[\mathcal{M}(\mathcal{\tilde{G}}, N, \beta) = \mathbf{\bar{g}}^*]$. 
Therefore, when $\mathcal{M}$ has a high dimensional output, it is very unlikely for the distribution of the algorithm's output to have a spike at any certain point (i.e. $\tilde{q} \ll 1$). Since the data-dependent privacy bound improves upon the data-independent bound only when $\tilde{q} \ll 1$, it is unlikely to benefit algorithms with high-dimensional output. Based on this understanding, we use 
\Cref{theorem:rdp_guarantee} (the data-independent privacy bound) for the privacy analysis in 
\sysname.
We also provide empirical evaluation of the data-dependent and data-independent privacy bounds in Figure~\ref{fig:dept_bound} in Section~\ref{sec:ablation}. 

%% file: convergence_analysis.tex
\textit{Why does top-$k$ and sign compression help the DP data generation process?} In this section,
we provide theoretical analysis on the convergence to present
the \textit{intuition} behind our proposed gradient compression and aggregation algorithm \top.
Note that, as directly analyzing 
the convergence of GAN is 
technically challenging~\citep{pmlr-v80-mescheder18a}
and beyond the scope of this paper,
we focus on an abstract 
model in which each teacher provides
an \textit{unbiased} gradient
estimator for SGD with loss function $F_n(x)$ given input $x$. We believe that
this is a plausible assumption since
in our setting each teacher has
access to a random non-overlapped partition of
the input data.

\m{
Understanding the convergence
behavior of stochastic gradient 
descent in the context of differential privacy is a challenging problem.
At the first glance, the DP noise 
might look like just another 
variance term over the stochastic gradient; however, it is the other
operations such as the \textbf{normalization}
and \textbf{clipping} of gradients
that make the analysis much harder.
In fact, it is not until recently~\citep{thakkar_adaptive_clipping2019, NEURIPS2020_gradient_clipping, adaclip2019} 
that researchers developed 
some results to analyze the behavior 
of DP-SGD with gradient \textit{norm} clipping (often limited to scaling $L^2$ norm instead of truncating).
In our context, this problem becomes
even more challenging, as we need to consider not only \textit{element-wise} gradient clipping, but also top-$K$ compression, an operator that introduces \textit{bias}, instead
of \textit{variance} to our
gradient estimator.}

\paragraph{Setup and Assumptions.}
We focus on the following setting in which our goal is to minimize $f(x) = \frac{1}{N}\sum_{n \in [N]} F_n(x)$ over $\mathbb{R}^d$. Recall that the update rule is \vspace{-1mm}
\begin{small}
\begin{equation}\label{eqn:update_rule}
x_{t+1} = x_t -  \frac{\gamma}{N} \sum_{n\in[N]} \left(Q ({\kk\texttt{clip}(\texttt{top-k}(F_{n}' (x_t)),c)}, \xi_t) + \mathcal{N} (0, Ak)\right),\vspace{-1mm}
\end{equation}
\end{small}for some constant $A>0$ \m{and a clipping constant $c>0$, with clipping performed \textit{coordinate-wise}}. 
Here we rephrased the stochastic sign quantization using $Q(x, \xi) =
\xi (x)$, when $x\geq 0$, and $Q(x, \xi) =-\xi (-x)$,  when $x <0$,
where $\xi (x) \sim Ber(x)$, element-wise. Thus the term $Q (\texttt{top-k}(F_{n}' (x_t)), \xi_t)$ is equivalent to  \Cref{algo:topk}, which takes in a gradient vector of a teacher model's gradient $F_{n}' (x_t)$ and returns the compressed gradient vector. Furthermore, $\mathcal{N} (0, Ak)$ is the noise
added to ensure differential privacy, since we know from \Cref{theorem:rdp_guarantee} that when \texttt{DPTopkAgg} satisfies $(\lambda, \alpha)$-RDP, the variance of Gaussian noise is $\sigma^2=2k\lambda/\alpha$, which is proportional to $k$.

Following previous work, we 
make a set of standard assumptions~\citep{NIPS2018_7837,10.5555/2969442.2969538,10.5555/3326943.3327063}.
We assume that $f$ has $L$-Lipschitz gradient, that all $F_n$ are smooth and that we have a bounded gradient, meaning that there exists $M>0$ such that $\frac{1}{N} \sum_{n\in [N]} \left\|F_n'(x)\right\|^2 \leq M^2$. \m{Furthermore, we assume bounded stochastic variance per coordinate, meaning that for every $i\in[d]$ there exists $\sigma_i>0$ such that $\frac{1}{N} \sum_{n\in[N]}|F_n'(x)-\nabla f(x)|^2 \leq \sigma_{i}^2$, for all $x\in \RR^d$.} With respect to compression, we see that $Q$ is unbiased
in our case, i.e. $\E_\xi \left[Q(x,\xi)\right] = x$, for all $x$, and of bounded variance, i.e. $\E_\xi \left[ \| Q(x,\xi) \right.$ $\left.  - x\|^2 \right] \leq \tilde{\sigma}^2$, for some $\tilde{\sigma} > 0$, and all $x$. Finally, with respect to \texttt{top-k}, we assume (see \citep{NIPS2018_7837}) that there exists a \m{non-increasing sequence $1 \geq \tau_1 \geq \ldots \geq \tau_d = 0$}, such that for all $k \in [d]$ and all $x \in \mathbb{R}^d$, one has 
$\| F_n'(x) - \texttt{top-k}(F_n'(x)) \| \leq  \tau_k \|F_n'(x)\|.
$
Given these assumptions, we have the following result:

\m{
\begin{theorem}\label{thmConvergenceAnalysis}
(Convergence of top-$k$ Mechanism with/without Gradient Quantization) Suppose that the above assumptions hold, and let $k\in [d]$. Then after $T$ updates using the learning rate $\gamma$, one has \vspace{-1mm}
\begin{small}
	\begin{align}\label{eqnMainThmEqn}
&\left(\frac{\min\{c,1\}}{d+2}\right)\frac{1}{T} \sum_{t\in [T]}  \min \{ \E \| \nabla f(x_t)\|^2,\E\|\nabla f(x_t)\|_1 \} \nonumber \\
&\leq \min \{\tau_k M^2, c(d-k)M\} + L\gamma A k + (f(x_0) - f(x^*))/(T\gamma) \nonumber \\
&\hspace{8mm}+\max \{\|\sigma\|^2 + \|\sigma\|M ,2 \|\sigma\|_1\} + 2L\gamma (\tilde \sigma^2 + \min \{c^2,M^2\}).
	\end{align}
\end{small}
Moreover, if no quantization is used, i.e. $Q(x,\xi)=x$ for all x, then one can improve the last term to $L\gamma \min \{c^2,M^2\}$.
\end{theorem}
}

\textbf{Proof Sketch.}
The full proof is given in~\Cref{sec:app_proofs}, whereas here we explain main ingredients. \m{Intuitively, clipping gradients yields a dichotomy between gradient performing as the usual gradient descent versus the signed gradient descent (as in \citep{signSGD}) of magnitude $c$.} We start with a well-known fact that $f$ having $L$-Lipschitz gradients implies $f(x_{t+1}) - f(x_t) \leq \inn{\nabla f(x_t)}{x_{t+1} - x_t} + \frac{L}{2} \|x_{t+1} - x_t\|^2$, which allows one to look at the convergence rate step by step. \m{Upon inserting the update rule~(\ref{eqn:update_rule}), we split the argument into two cases based on, for $i\in[d]$ and \begin{small}$A_i:=\{n\in[N]\colon |F_n'(x)|\geq c\}$\end{small},\vspace{-0.5mm}
\begin{small}
 \begin{align*}
	&\texttt{clip}(F_{n}'(x)_i,c) = c \cdot  \texttt{sign} (F_{n}'(x)_i) \cdot \mathbf{1} \{n\in A_i\} + F_{n}'(x) \cdot \mathbf{1} \{n\notin A_i\}.
 \end{align*}\vspace{-0.5mm}
\end{small}Using a proof by contradiction, we show that the error terms cannot beat the main term for clipped and non-clipped gradients simultaneously.} In doing so, the error terms on the RHS of~(\ref{eqnMainThmEqn}) originate from the following: $\min\{\tau_{k}M^2,{c(d-k)M}\}$ comes from applying the top-$k$ mechanism \m{on top of clipped gradients}, $2L\gamma A k$ originates from the variance of the noise attributed to differential privacy, ${f_{0,*}}/{T\gamma}$ comes from the telescoping property when summing over all steps. \m{The term $\max \{\|\sigma\|^2 + \|\sigma\|M ,2 \|\sigma\|_1\}$ comes from the clipping dichotomy (also contributing to the term $\min\{c,1\}$ on the LHS),} whereas $2L\gamma (\tilde{\sigma}^2 + \min\{c^2,M^2\})$ is the variance of quantization step. The without quantization case follows the similar approach, up to the non-existence of randomness in the quantization case, yielding a simpler proof.

\textbf{Discussion: Why Does Top-K Help?} \label{sec:discussion}
The above result depicts the following
tradeoff.
As $k$ gets \emph{smaller} the 
error caused by top-$k$ quantization gets larger,
leading to two effects:
\begin{enumerate}
\item The term 
\m{$\min \{\tau_k M^2, c(d-k)M\}$}, introduced through the \textit{bias} of top-$k$ compression,
gets larger;  
\item The $2L\gamma Ak$ term, introduced by the differential privacy noise, however, gets smaller.
\end{enumerate}

\m{Given a finite number of iterations $T$, in the worst case the bias introduced through the term $\tau_{k}$ dominates when the gradients are evenly distributed over coordinates, yielding that the top-$k$ compression can significantly slow down the convergence rate in the worst case. However, previous works~\citep{NIPS2018_7837,10.5555/3326943.3327063}  empirically verify that under certain real distribution of gradient dimensions, the top-$k$ compression does not introduce a large bias, yielding justification for top-$k$ compression, especially when the original dimension $d$ is of very high dimension.} For example, if we 
assume that the gradient follows the \textit{Weibull distribution} $W(\rho_1, \rho_2)$, for some $\rho_1 > 0$ and $0<\rho_2 < 1$, following recent work in gradient compression~\citep{tinyscript}, then $\tau_k$ are, on expectation, distributed as $\tau_k \propto \exp\left(-(k/\rho_1 d)^{\rho_2}\right) - \exp(-1)$, which for small $\rho_2$ grows significantly slower than the contribution of the noise due to differential privacy (linear in $k$) decreases, as $k$ decreases. 
\m{Thus, the convergence-privacy tradeoff for algorithm \top can be clearly characterized. It is obvious that given the convergence guarantee, the compression step could 
save the privacy budget and therefore improve the utility (i.e. smaller DP noise is added) for training on high-dimensional data, as long as the chosen $k$ is not too small. }

\subsection{Discussion: \top for SGD Training}
In addition to the DP generative model, the proposed DP gradient compression and aggregation algorithm \top, which is a key building block of \sysname, is also generalizable for the standard DP SGD training by applying the gradient compression and aggregation in the DP SGD training process. 
However, since the DP SGD algorithm has already achieved high data utility, the improvement with \top is empirically marginal, and \m{we will defer the details on  how to adapt \top to training a differentially private deep neural network and the corresponding evaluation in~\Cref{sec:appen-top}.}

%% file: tables/tab_main.tex
\begin{table*}[t!]
\centering
\caption{\small \textbf{Performance of different differentially private data generative models on Image Datasets:} 
Classification accuracy of the model trained on the generated data and tested on real test data under different $\eps$ ($\delta=10^{-5}$). 
%
}
\resizebox{.8\linewidth}{!}{%
\begin{tabular}{l|c|c|ccccc}
\toprule
\diagbox[width=10em]{\textbf{Dataset}}{\textbf{Methods}} 
&\textbf{DC-GAN ($\eps=\infty$)} & $\varepsilon$ &\textbf{DP-GAN} & \textbf{PATE-GAN}  &\textbf{G-PATE} & \m{\textbf{GS-WGAN}} &\textbf{\name{}}  \\ \midrule
\multirow{2}{*}{\textbf{MNIST}} & \multirow{2}{*}{\shortstack{0.9653 }} & $\varepsilon=1$& 0.4036
& 0.4168 & {0.5810} & \m{0.1432} & \textbf{0.7123}  \\ 
&   & $\varepsilon=10$ & {0.8011} &  0.6667 &\textbf{0.8092} & \m{0.8075} & {0.8066}  \\ \midrule
\multirow{2}{*}{\textbf{{Fashion-MNIST}}} & \multirow{2}{*}{\shortstack{0.8032 }}& $\varepsilon=1$ &0.1053 & 0.4222& {0.5567} & \m{0.1661} &  \textbf{0.6478} \\ 
& & $\varepsilon=10$ & 0.6098  & 0.6218 & {0.6934} & \m{0.6579} & \textbf{0.7061}  \\  \midrule
\multirow{2}{*}{\textbf{{CelebA-Gender}}} & \multirow{2}{*}{\shortstack{0.8149 }} & $\varepsilon=1$& 0.5330 & 0.6068 & {0.6702} & \m{0.5901}  & \textbf{0.7058} \\ 
&   & $\varepsilon=10$ & 0.5211 & {0.6535} & {0.6897} & \m{0.6136} & \textbf{0.7287} \\ \midrule
\multirow{2}{*}{\textbf{{CelebA-Hair}}} & \multirow{2}{*}{\shortstack{0.7678 }}& $\varepsilon=1$ & 0.3447 & 0.3789& {0.4985} & \m{0.4203} & \textbf{0.6061} \\ 
& & $\varepsilon=10$ & 0.3920 & 0.3900 & {0.6217} &  \m{0.5225} & \textbf{0.6224}  \\ 
\midrule
\m{\multirow{2}{*}{\textbf{{Places365}}}} & \m{\multirow{2}{*}{\shortstack{0.7404}}} & \m{$\varepsilon=1$} & \m{0.3200} & \m{0.3238}& \m{0.3483} & \m{0.3375} & \m{\textbf{0.4313}} \\ 
& & \m{$\varepsilon=10$} & \m{0.3292} & \m{0.3796} & \m{0.3883} & \m{0.3725} & \m{\textbf{0.4875}}  \\ 

\bottomrule
\end{tabular}%
}
\label{tab:image}
\vspace{-1em}
\end{table*}

%% file: tables/tab_main_small.tex
\begin{table*}[th]
\centering
\caption{\small \textbf{Performance Comparison of different differentially private data generative models on Image Datasets under small privacy budget which provides strong privacy guarantees}  ($\eps \leq 1$, $\delta=10^{-5}$). }
\label{tab:lowdp}
\resizebox{.9\linewidth}{!}{%
\begin{tabular}{c|ccccc|ccccc}
\toprule
\multirow{2}{*}{\textbf{$\varepsilon$}} & \multicolumn{5}{c|}{\textbf{MNIST}} &  \multicolumn{5}{c}{\textbf{Fashion-MNIST}} \\ 
 & DP-GAN & PATE-GAN  & G-PATE & \m{GS-WGAN} & \name{} & DP-GAN & PATE-GAN  & G-PATE & \m{GS-WGAN} & \name{} \\
\midrule
0.2 & 0.1104 & 0.2176 & {0.2230} & \m{0.0972}  & \textbf{0.2344} & 0.1021  & 0.1605 & 0.1874     &  \m{0.1000} & \textbf{0.2226} \\
0.4 & 0.1524 & 0.2399 & {0.2478} & \m{0.1029}  & \textbf{0.2919} & 0.1302  & 0.2977 & 0.3020     &  \m{0.1001} & \textbf{0.3863} \\
0.6 & 0.1022 & 0.3484 & 0.4184   & \m{0.1044} & \textbf{0.4201} & 0.0998    & 0.3698 & 0.4283    &  \m{0.1144} & \textbf{0.4314} \\
0.8 & 0.3732 & 0.3571 & 0.5377   & \m{0.1170}    & \textbf{0.6485} & 0.1210    & 0.3659 & 0.5258 &  \m{0.1242} & \textbf{0.5534} \\
1.0 & 0.4046 & 0.4168 & 0.5810   & \m{0.1432}  & \textbf{0.7123} & 0.1053    & 0.4222 & 0.5567   &  \m{0.1661} & \textbf{0.6478} \\
\bottomrule
\end{tabular}
}
\vspace{-1em}
\end{table*}

%% file: tables/tab_visual.tex
\begin{table}[t!]
\centering
\caption{\small \m{Quality evaluation of images generated by different differentially private data generative models on Image Datasets: 
we use Inception Score (IS) to measure the visual quality of the generated data  under different $\eps$ ($\delta=10^{-5}$).} %
}
\label{tab:visual}
\centering
\m{
\resizebox{\linewidth}{!}{%
\begin{tabular}{l|c|c|C{0.7cm}C{0.7cm}C{0.7cm}C{0.7cm}c}
\toprule
{\footnotesize\textbf{Dataset}} & \textbf{\scriptsize\makecell{Real\\ data}}  & $\bm\varepsilon$ & {\scriptsize\textbf{\makecell{DP-\\GAN}}} & {\scriptsize\textbf{\makecell{PATE-\\GAN}}}  & {\scriptsize\textbf{\makecell{G-\\PATE}}} & {\scriptsize\textbf{\makecell{GS-\\WGAN}}} &{\scriptsize\textbf{\name{}}}  \\ \midrule
\multirow{2}{*}{{\footnotesize\textbf{MNIST}}} & \multirow{2}{*}{9.86} &  $1$ & 1.00 & 1.19  & 3.60 & 1.00 & \textbf{{4.37}}  \\ 
                                &                       & $10$ &  1.00 & 1.46 & 5.16 &  \textbf{8.59} & 5.78 \\ \midrule
\multirow{2}{*}{{\footnotesize\textbf{\makecell{Fashion-\\MNIST}}}} & \multirow{2}{*}{9.01} & $1$ &  1.03 & 1.69 & 3.41 & 1.00 & \textbf{{3.93}}  \\ 
                                        &                       & $10$ & 1.05 & 2.35 & 4.33 & \textbf{5.87}  & 4.58 \\ \midrule
\multirow{2}{*}{\footnotesize\textbf{CelebA}} & \multirow{2}{*}{1.88} &   $1$ &  1.00 & 1.15 & 1.11 & 1.00 & \textbf{{1.18}}  \\ 
                                 &                       &  $10$ & 1.00 & 1.16 & 1.12 & 1.00 & \textbf{{1.42}}   \\ 
\bottomrule
\end{tabular}
}
}
\vspace{-0.5em}
\end{table}

%% file: tables/fig_dp_bound_main.tex
\begin{figure}[t]
    \centering
    \begin{subfigure}[t]{.48\columnwidth}
        \centering
        \includegraphics[width=\linewidth]{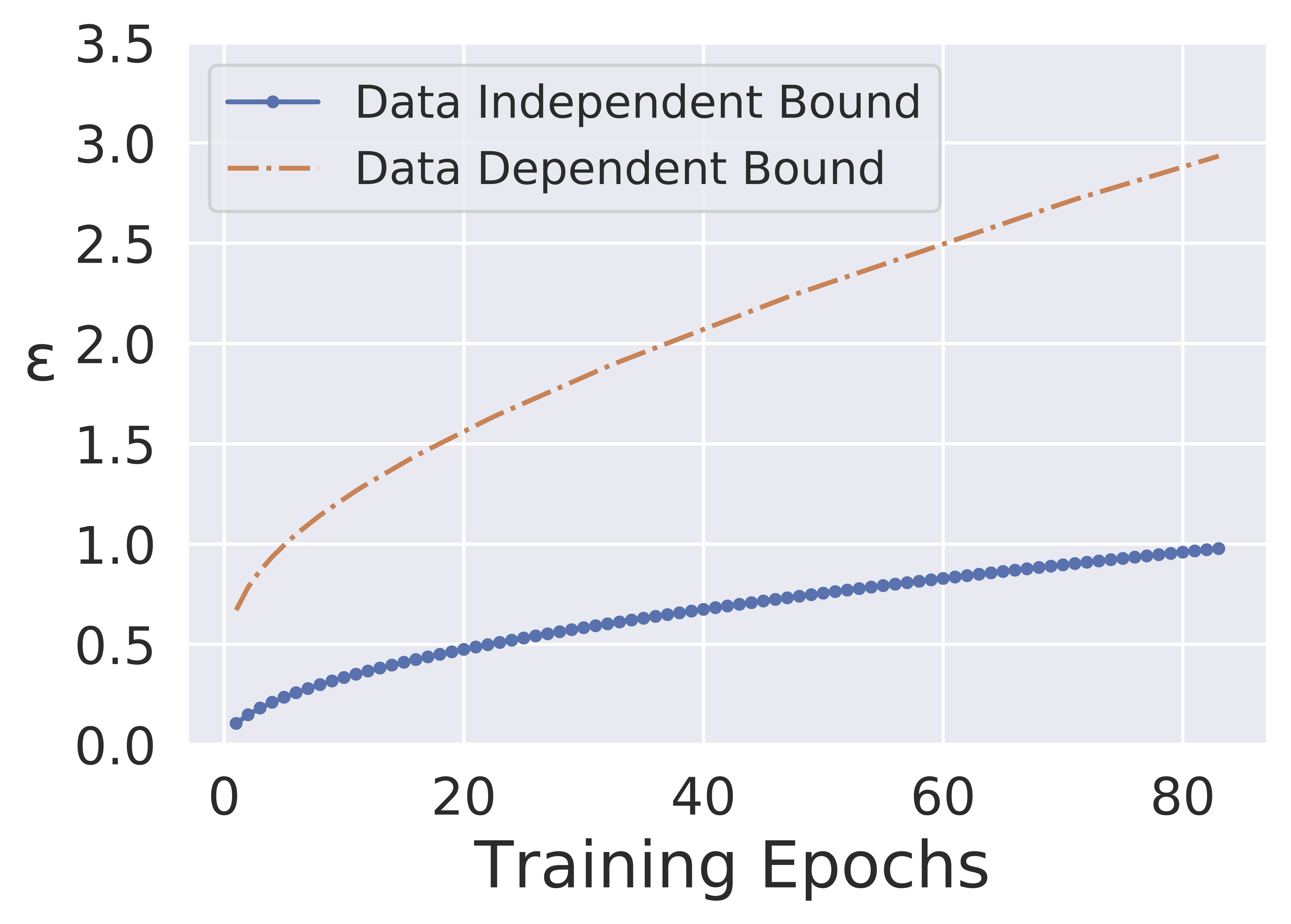}
    \end{subfigure}
    \begin{subfigure}[t]{.48\columnwidth}
        \centering
        \includegraphics[width=\linewidth]{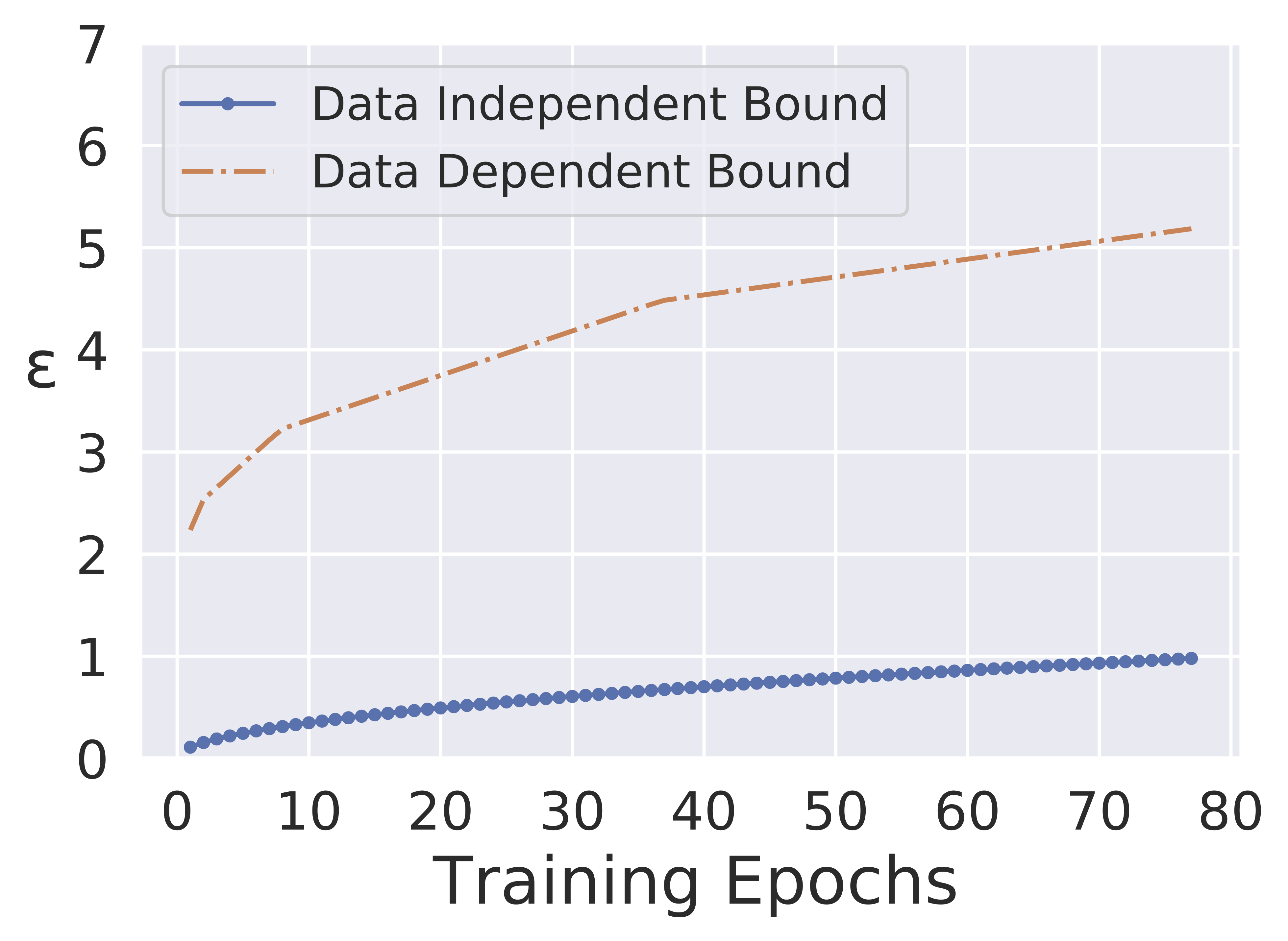}
    \end{subfigure}
    \vspace{-3mm}
    \caption{\small \m{Ablation studies on the data dependent bound v.s. data independent bound on MNIST (left) and CelebA-Hair (right).
    The data independent bound always yields tighter privacy bound than the data dependent analysis, given high dimensionality of  gradients.}
}
    \label{fig:dept_bound}
\end{figure}

%% file: tables/tab_hyperpara_all.tex
\begin{table*}[t]
\centering
\caption{\small \textbf{Impact of different hyper-parameters: the number of Teachers, top-$k$ and threshold $\beta$ under $\eps=1$ and $\delta=10^{-5}$.} We search for the optimal parameter combinations and report the best accuracy by controlling the parameter in each cell.
} 
\label{tab:hyperparam}
\begin{subtable}[]{0.48\linewidth}
\label{tab:projection_teachers_12}
\caption{Hyper-parameters Search for MNIST and Fashion-MNIST}
\resizebox{\columnwidth}{!}{%
\begin{tabular}{c|ccc|ccc}
\toprule
\multirow{2}{*}{}  & \multicolumn{3}{c|}{\textbf{Top-$k$}} & \multicolumn{3}{c}{\textbf{\# of Teachers}} \\ 
&  100 &  {200} & 300  & 2000  & 3000  & {4000}  \\ \midrule
\textbf{MNIST} & 0.5889  & \textbf{0.7123}  & 0.6753  & 0.5841 & 0.7061  & \textbf{0.7123}   \\ 
\textbf{Fashion} & 0.5738  & \textbf{0.6478} & 0.6088 & 0.5608 & 0.5952 & \textbf{0.6478}  \\ \bottomrule
\toprule
$\beta$ &  0 & 0.1 & \multicolumn{1}{c}{0.3} & 0.5 & 0.7 & 0.9  \\ \midrule
\textbf{MNIST} & 0.6361 & 0.6450  & \multicolumn{1}{c}{0.6890} & 0.6921 & \textbf{0.7123} & 0.6956   \\ 
\textbf{Fashion} & 0.5859 & 0.6103 & \multicolumn{1}{c}{0.6060}  & 0.6122 & 0.6213 & \textbf{0.6478}  \\ 
\bottomrule
\end{tabular}
}
\end{subtable}
\hfill
\begin{subtable}[]{0.48\linewidth}
\label{tab:projection_teachers_22}
\caption{Hyper-parameters Search for CelebA-Hair and CelebA-Gender}
\resizebox{\columnwidth}{!}{%
\begin{tabular}{c|ccc|ccc}
\toprule
\multirow{2}{*}{}  & \multicolumn{3}{c|}{\textbf{Top-$k$}} & \multicolumn{3}{c}{\textbf{\# of Teachers}} \\ 
&  500 & {700}  & 900  &  4000  & 6000  & 8000  \\ \midrule
\textbf{CelebA-Gender} & 0.6922 & \textbf{0.7058}  &  0.6811  &  0.6378  &  \textbf{0.7058}  & 0.6936   \\ 
\textbf{CelebA-Hair} & 0.5792   &  \textbf{0.6061} &  0.5769  & 0.5669 &  0.5835 &  \textbf{0.6061} \\ \bottomrule
\toprule
$\beta$ &  0.5  & 0.6 & \multicolumn{1}{c}{0.7}  & 0.8 & 0.85 & 0.9  \\ \midrule
\textbf{CelebA-Gender} & 0.6440   &  0.6789 & \multicolumn{1}{c}{0.6922} & 0.6861 &  \textbf{0.7058} & 0.6381 \\ 
\textbf{CelebA-Hair} &  0.4957 &  0.5669 & \multicolumn{1}{c}{0.5612}  & {0.6022}  & 0.5835 & \textbf{0.6061} \\ 
\bottomrule
\end{tabular}%
}
\end{subtable}
\vspace{-0.8em}
\end{table*}

%% file: tables/tab_compression.tex
\begin{table}[t]
\centering
\caption{\small \m{\textbf{Accuracy Comparison of different gradient compression methods (\top, D$^2$P-\textsc{Fed}, FetchSGD).}  We report the test classification accuracy of models trained with data generated with each technique under $\eps=1$ and $\delta=10^{-5}$.}}\label{tab:compression}
\vspace{-2mm}
\tabcolsep=0.11cm
\resizebox{0.8\columnwidth}{!}{
\m{
\begin{tabular}{l|ccc}
\toprule
\diagbox[width=10em]{\textbf{Dataset}}{\textbf{Methods}} & {\textbf{\top}}  & \textbf{D$^2$P-\textsc{Fed}}  & {\textbf{FetchSGD}}     \\
\midrule
{\textbf{MNIST}} & \textbf{0.7123} & {0.1424} & 0.6935   \\ 
\midrule
{\textbf{Fashion-MNIST}} & \textbf{0.6478} & {0.1667} & 0.6387   \\ 
\midrule
{\textbf{CelebA-Gender}} & \textbf{0.7058}  & 0.4445 &  0.6552  \\ 
\midrule
{\textbf{CelebA-Hair}} & \textbf{0.6061} & 0.2893 & 0.4926   \\ 
\bottomrule
\end{tabular}%
}
}
\end{table}

%% file: tables/tab_components.tex
\begin{table}[t]
\centering
\caption{\small \m{\textbf{Ablation studies on the impact of different components of \sysname pipeline on Image Datasets:}  We report the test classification accuracy of models trained with data generated
based on different variants of \sysname
under $\eps=1$, $\delta=10^{-5}$. The first row of each data groups presents the performance of \sysname.}}
\vspace{-2mm}
\resizebox{\columnwidth}{!}{\m{
\begin{tabular}{l|C{1.6cm}C{1.6cm}C{1.6cm}|c}
\toprule
\multirow{2}{*}{\diagbox[width=10em]{\textbf{Dataset}}{\textbf{Component}}} & \multirow{2}{*}{\textbf{\small Top-$k$}}  & \textbf{\small Stochastic}  & {\textbf{\small Aggregation}}   &  \multirow{2}{*}{\textbf{\small Accuracy}} \\ 
& & \textbf{\small Quantization} & \textbf{\small Thresholding} &  \\
\midrule
\multirow{4}{*}{\textbf{MNIST}} & \cmark & \cmark & \cmark &\textbf{0.7123}  \\ 
\cmidrule{2-5}
&  \xmark  & \cmark  & \cmark  & 0.5170 \\
& \cmark & \xmark  & \cmark  &  0.6741 \\
& \cmark & \cmark & \xmark  & 0.6361 \\
\midrule
\multirow{4}{*}{\makecell{\textbf{Fashion-}\textbf{MNIST}}} & \cmark & \cmark & \cmark & \textbf{0.6478}  \\ 
\cmidrule{2-5}
& \xmark  & \cmark  & \cmark & 0.4775 \\
& \cmark & \xmark  & \cmark &  0.6159 \\
& \cmark & \cmark & \xmark  & 0.5859 \\
\midrule
\multirow{4}{*}{\makecell{\textbf{CelebA-}\textbf{Gender}}} & \cmark & \cmark & \cmark & \textbf{0.7058}  \\ 
\cmidrule{2-5}
& \xmark  & \cmark  & \cmark & 0.6134 \\
& \cmark & \xmark   & \cmark &  0.6889 \\
& \cmark & \cmark & \xmark  &  0.6860 \\
\midrule
\multirow{4}{*}{\makecell{\textbf{CelebA-}\textbf{Hair}}} & \cmark & \cmark & \cmark & \textbf{0.6061}  \\ 
\cmidrule{2-5}
& \xmark  & \cmark  & \cmark &  0.3318 \\
& \cmark & \xmark  & \cmark &  0.5325 \\
& \cmark & \cmark & \xmark  &   0.5504 \\
\bottomrule
\end{tabular}%
}
}
\label{tab:components}
\end{table}

%% file: related_work.tex

\textit{DP Generative Models.} 
In order to generate data with differential privacy guarantees, several works have been conducted to develop DP generative models for low-dimensional data such as tabular data. 
Some of them apply differential privacy to traditional data generation algorithms, such as Bayesian networks~\citep{zhang2017privbayes}, synthetic data generation from marginal distributions~\citep{qardaji2014priview}, and the multiplicative weights approach~\citep{hardt2012simple}. Although these methods have demonstrated good performances on low dimensional datasets, they suffer from either low data utility or high sampling complexity on high dimensional data, and therefore they are usually not suitable for the high-dimensional image datasets discussed in this paper. 

Another line of work 
adapts DP-SGD to GAN. DPGAN~\citep{xie2018differentially} achieves differential privacy by adding Gaussian noise to the discriminator gradients during the training process. DP-CGAN~\citep{torkzadehmahani2019dp} uses a similar approach to guarantee DP and trains a conditional GAN to generate both synthetic data and labels. GS-WGAN~\citep{chen2020gs} uses the Wasserstein loss and sanitizes the data-dependent gradients of the generator to improve data utility. However, these approaches still suffer from low data utility when applied to high dimensional datasets due to privacy budget explosion. 

PATE-GAN~\citep{yoon2018pategan} combines the PATE framework with GAN. It trains multiple teacher discriminators and uses them to update the student discriminator. However, in this framework, it essentially applies PATE to train the discriminator within a GAN.
Both the teacher and students models are discriminators and the interaction between the generator and discriminator is not adapted for the  teacher-student framework. Thus, PATE-GAN is also only evaluated on low dimensional tabular data and suffers the similar problem under limited privacy budget.
G-PATE~\citep{long2019scalable} improves upon PATE-GAN by directly training a student generator using the teacher discriminators. It uses the random projection algorithm to reduce the gradient dimension during training, which is challenging to analyze its convergence. 
By combining the PATE framework with top-$k$ gradient compression, \sysname demonstrates a significant utility improvement upon PATE-GAN and G-PATE on high dimensional datasets, with theoretical analysis on its convergence. 

\mo{
\textit{DP SGD Training}. 
DPDL~\citep{Abadi2016DeepLW} is the first work that applies the notion of Differential Privacy to the SGD training to prevent deep neural models from exposing private information of training data. DPDL also proposes to compute the privacy cost of the training by moments accountant, which proves to be a tighter bound than the strong composition theorem. \citet{mcmahan2019tfprivacy} adopts the notion of R\'enyi differential privacy, which extends and generalizes the moment accountant to multi-vector queries. Thus, the R\'enyi differential privacy analysis enables the framework to provide privacy for heterogeneous sets of vectors and is widely adopted by current open-source DP library (Tensorflow Privacy and Pytorch Opacus) implementation.
However, the high-dimensional data issue is still present in these algorithms given the fact that the privacy budget can be consumed quickly when aggregating these gradients in a differentially private manner.
}

\textit{Gradient Compression.}
Communication efficient distributed learning has attracted
intensive interests recently. Popular
techniques include 
gradient compression~\citep{NIPS2017_6768,NIPS2018_7837,10.5555/2969442.2969538,10.5555/3326943.3327063}, decentralization~\citep{pmlr-v97-koloskova19a, NIPS2017_7117},
and asynchronization~\citep{NIPS2015_5751} (see \citep{10.1145/3320060}).
The essence of these methods
is to reason about the  \textit{noise} 
introduced via relaxations
in the system design. 
cpSGD~\citep{NIPS2018_7984} is proposed as a binomial DP-mechanism specifically designed 
for \m{stochastic $k$-level} gradient quantization\m{~\citep{mcmahan2017communication}}
to allow low-precision 
communication after
adding DP noises. 
\m{Extending this work, D$^2$P-\textsc{Fed}~\citep{wang2020d2p} instead applies the discrete Gaussian mechanism to the same $k$-level quantization and achieves a stronger privacy guarantee.} 
\m{Similarly, Kairouz et al.~\citep{kairouz2021distributed} combine discrete Gaussian mechanism with $k$-level quantization to facilitate federated learning with differential privacy and secure aggregation.}  
\m{In comparison, \sysname uses PATE framework to give rigorous privacy guarantee and apply sign compression as teacher voting to save privacy budget. }
\m{FetchSGD~\citep{rothchild2020fetchsgd} focuses on communication-efficiency in the federated learning setting, and proposes Count Sketch data structure and top-$k$ operation for fast gradient compression and aggregation. However, FetchSGD lacks the discussion for privacy guarantee.}
In this paper,  we explore the
 relationship between privacy and gradient compression in 
a different scenario and 
illustrate how 
gradient compression can 
help to achieve 
better utility in 
privacy preserving algorithms
over high dimensional data.
\m{We propose \top by combining stochastic sign~\citep{jin2020stochastic} with top-$k$ gradient compression. Our empirical results show that \top outperforms state-of-art gradient compression algorithms on improving model utility with differential privacy guarantee.}

%% file: conclusions.tex
\section{Conclusion}

Overall, we propose a novel and effective differentially private data generative model \sysname, which is applicable to high-dimensional data compared with existing approaches. In addition, we propose a novel algorithm \top to perform gradient compression  and aggregation. We provide the DP analysis as well as convergence analysis for the proposed model.  Extensive empirical experiments demonstrate that \sysname  substantially outperforms the existing DP generative models on different especially high-dimensional image datasets, even under limited privacy budget.

\section*{Acknowledgement}
We thank the anonymous reviewers for their constructive feedback. We also thank Dingfan Chen and many others for the helpful discussion. This work is partially supported by the NSF grant No.1910100, NSF CNS 20-46726 CAR, and Amazon Research Award.

%% file: tables/tab_control.tex
\begin{table}[t]
    \centering
    \caption{\small Results of the control experiments to explore the impact of gradient compression and noise injection. }
    \label{tab:control}
    \begin{subtable}[]{\columnwidth}
    	\centering
    	\caption{\small Experimental setup. Each cell is one experimental scenario.}
        \label{tab:control-setup}
    	\resizebox{\columnwidth}{!}{%
            \begin{tabular}{c|ccc}
            \toprule
                 & noise $\sim {\cal N}(0,\sigma^2 C^2\bf I)$ & noise $\sim{\cal N}(0,k\sigma^2 C^2\bf I)$ & no noise \\\midrule
            \texttt{NormTopK} &  TopK-GM-DP & \top & TopK-SGD \\
            no compression & \gmdp~\citep{mcmahan2019tfprivacy} & -- & \makecell{clipped SGD}\\\bottomrule
            \end{tabular}
	    }
	    \vspace{1em}
    \end{subtable}
    \begin{subtable}[]{\columnwidth}
    	\centering
    	\caption{\small Experimental results on MNIST dataset under small and big privacy budgets. (above) $\eps=0.1$ and $k=0.8$; (below) $\eps=1.0$ and $k=0.8$.}
        \label{tab:control-res-1}
    	\resizebox{\columnwidth}{!}{%
            \begin{tabular}{c|ccc}
            \toprule
                 & noise $\sim{\cal N}(0,\sigma^2 C^2\bf I)$ & noise $\sim{\cal N}(0,k\sigma^2 C^2\bf I)$ & no noise \\\midrule
            \texttt{NormTopK} & 83.86 $\pm$ {\small 1.49}  & 85.88 $\pm$ {\small 1.97} & 91.96 $\pm$ {\small 0.61} \\
            no compression & 85.05 $\pm$ {\small 1.85} & -- & 94.26 $\pm$ {\small 0.43}\\\bottomrule
            \end{tabular}%
	    }
    	\resizebox{\columnwidth}{!}{%
    	        \begin{tabular}{c|ccc}
            \toprule
                 & noise $\sim{\cal N}(0,\sigma^2 C^2\bf I)$ & noise $\sim{\cal N}(0,k\sigma^2 C^2\bf I)$ & no noise \\\midrule
            \texttt{NormTopK} & 94.95 $\pm$ {\small 0.25}  & 95.42 $\pm$ {\small 0.20} & 98.03 $\pm$ {\small 0.09} \\
            no compression & 95.56 $\pm$ {\small 0.17} & -- & 98.94 $\pm$ {\small 0.04}\\\bottomrule
        \end{tabular}
	    }
    \end{subtable}
    \label{tab:control-res}
\end{table}